\documentclass[twocolumn, final]{asme2e}

\newcommand{\be}{\begin{equation}}
\newcommand{\ee}{\end{equation}}
\newcommand{\beq}{\begin{equation}}
\newcommand{\eeq}{\end{equation}}
\newcommand{\bed}{\begin{displaymath}}
\newcommand{\eed}{\end{displaymath}}
\newcommand{\beqa}{\begin{eqnarray}}
\newcommand{\eeqa}{\end{eqnarray}}
\newcommand{\beqann}{\begin{eqnarray*}}
\newcommand{\eeqann}{\end{eqnarray*}}
\newcommand{\bseq}{\begin{subequations}}
\newcommand{\eseq}{\end{subequations}}

\newcommand{\ba}{\begin{array}}
\newcommand{\ea}{\end{array}}

\newcommand{\negr}[1]{{\bf {#1}}}

\usepackage{amsbsy}
\usepackage[dvips]{graphicx}
\usepackage{psfrag}
\usepackage{epsf}
\usepackage{epsfig}
\usepackage{array}
\usepackage{amssymb}
\usepackage{subfigure}
\usepackage{subeqn}
\def\confshortname{DETC'2004}
\def\conffullname{2004 ASME Design Engineering Technical Conferences}
\def\confdate{28}
\def\confmonth{September 28-October 2}
\def\confyear{2004}
\def\confcity{Salt Lake City}
\def\confstate{Utah}
\def\confcountry{USA}

\def\papernum{DETC2004/57168}

\title{The Design of a Novel Prismatic Drive \goodbreak for a Three-DOF Parallel-Kinematics Machine}
\author{J. Renotte, D. Chablat
    \affiliation{
      Institut de Recherche en Communications \\et Cybern\'etique de
      Nantes, UMR CNRS n$^\circ$ 6597 \\
      1, rue de la No\"e, 44321 Nantes, France \\
      Damien.Chablat@irccyn.ec-nantes.fr
    }}
\author{J. Angeles
    \affiliation{
      Department of Mechanical Engineering \& \\
      Centre for Intelligent Machines, McGill University \\
      817 Sherbrooke Street West, Montreal, Canada H3A 2K6 \\
      angeles@cim.mcgill.ca
    }}
\begin{document}
\maketitle
\begin{abstract}
The design of a novel prismatic drive is reported in this paper.
This transmission is based on {\it Slide-O-Cam}, a cam mechanism
with multiple rollers mounted on a  common translating follower.
The design of Slide-O-Cam was reported elsewhere. This drive thus
provides pure-rolling motion, thereby reducing the friction of
rack-and-pinions and linear drives. Such properties can be used to
design new transmissions for parallel-kinematics machines. In this
paper, this transmission is optimized to replace ball-screws in
Orthoglide, a three-DOF parallel robot optimized for machining
applications.
\end{abstract}
\section{Introduction}
In robotics and mechatronics applications, whereby motion is
controlled using a piece of software, the conversion of motion
from rotational to translational is usually done by {\em ball
screws} or {\em linear actuators}. Of these alternatives, ball
screws are gaining popularity, one of their drawbacks being the
high number of moving parts that they comprise, for their
functioning relies on a number of balls rolling on grooves
machined on a shaft; one more drawback of ball screws is their low
load-carrying capacity, stemming from the punctual form of contact
by means of which loads are transmitted. Linear bearings solve
these drawbacks to some extent, for they can be fabricated with
roller bearings, their drawback being that these devices rely on a
form of direct-drive motor, which makes them expensive to produce
and to maintain. A novel transmission, called {\it Slide-O-Cam},
was introduced in \cite{Gonzalez-Palacios:2000}
(Fig.~\ref{fig001}) to transform a rotation into a translation.
Slide-O-Cam is composed of four major elements: (i) the frame,
(ii) the cam, (iii) the follower and (iv) the rollers. The input
axis on which the cam is mounted, the camshaft, is driven at a
constant angular velocity. Power is transmitted to the output, the
translating follower, which is the roller-carrying slider, by
means of pure-rolling contact between cam and roller. The roller
comprises two components, the pin and the bearing. The bearing is
mounted at one end of the pin, while the other end is press-fit
into the roller-carrying slider. Contact between cam and roller
thus takes place at the outer surface of the bearing. The
mechanism uses two conjugate cam-follower pairs, which alternately
take over the motion transmission to ensure a positive action;
rollers are driven by the cams, throughout a complete cycle. The
main advantage of using a cam-follower mechanism instead of an
alternative transmission to transform rotation into translation is
that contact through a roller reduces friction, contact stress and
wear.

This transmission will be optimized to replace the three ball
screws used by the Orthoglide prototype \cite{Chablat:2003}.
Orthoglide features three prismatic joints mounted orthogonaly,
three identical legs and a mobile platform, which moves in the
Cartesian $x$-$y$-$z$ space with fixed orientation, as shown in
Fig.~\ref{fig002}. The motor used to move each axis is SANYO DENKI
(ref. P30B08075D) with a constant torque of 1.2~Nm from 0 to
3000~rpm. This property enables the mechanism to move throughout
the workspace a 4~kg load with an acceleration of 17~ms$^{-2}$ and
a velocity of 1.3~ms$^{-1}$. On the ball screws, the pitch is
50~mm per cam turn. The minimum radius of the camshaft is 8.5~mm.
{\begin{figure}[htb]
 \begin{minipage}[b]{4.4cm}
 \begin{center}
   {\scriptsize
   \psfrag{Roller}{Roller}
   \psfrag{Follower}{Follower}
   \psfrag{Conjugate}{Conjugate}}
   \centerline{\epsfig{file = 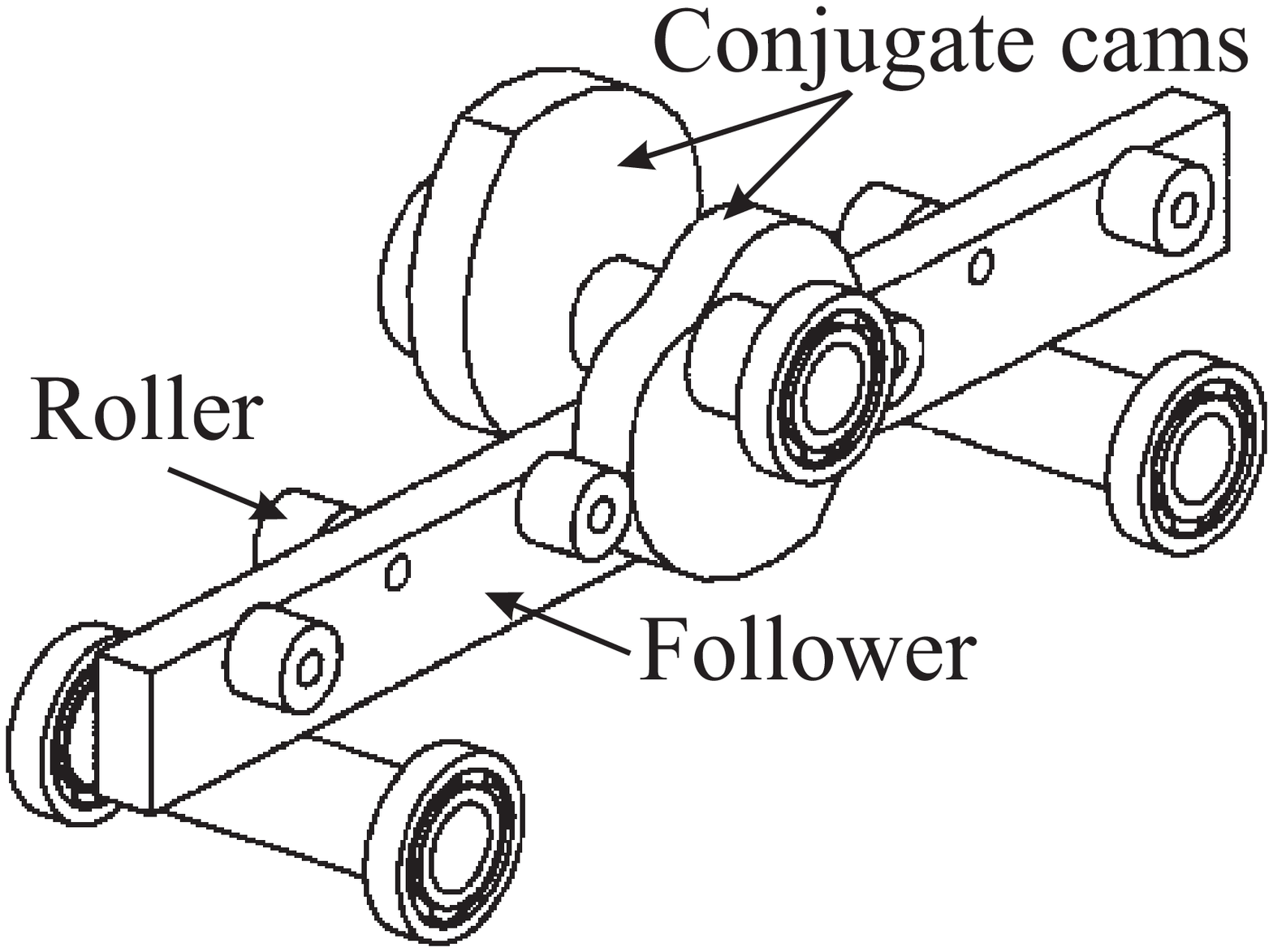,scale = 0.2}}
   \caption{Layout of \goodbreak Slide-O-Cam}
   \label{fig001}
 \end{center}
 \end{minipage}
 \begin{minipage}[b]{4.4cm}
 \begin{center}
    \centerline{\epsfig{file = 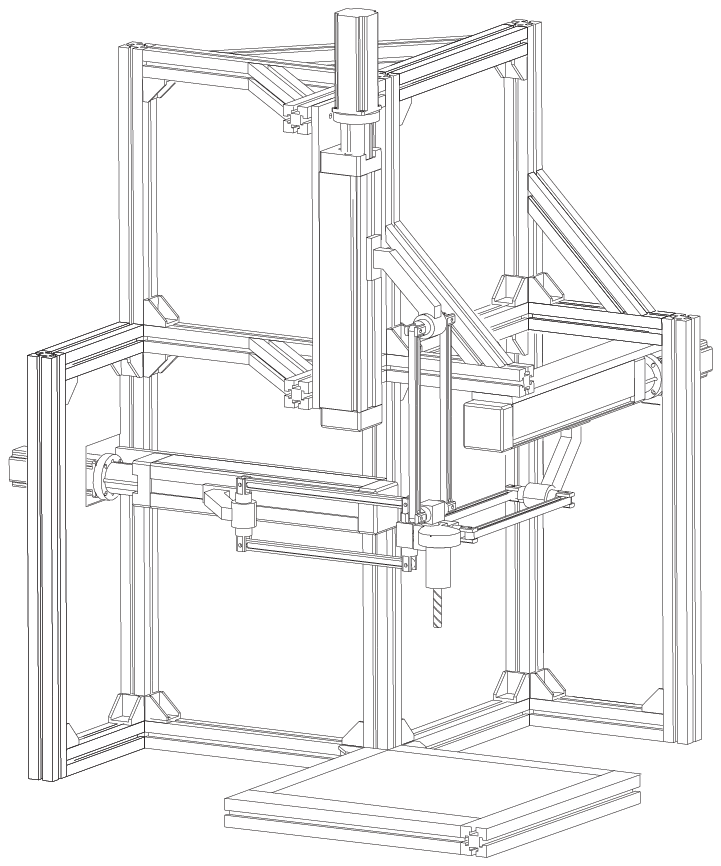,scale = 0.35}}
    \caption{The Orthoglide}
    \label{fig002}
 \end{center}
 \end{minipage}
 {\vspace{-1cm}}
 \end{figure}
Unlike Lampinen \cite{Lampinen:2003}, who used a genetic
algorithm, we use a deterministic method, while taking into
account geometric and machining constraints as outlined in
\cite{Bouzakis:1997}. In section 2, we introduce the relations
describing the cam profile and the mechanism kinematics. In
Section~3, we derive conditions on the design parameters so as to
have a fully convex cam profile, to avoid \textit{undercutting},
and to have a geometrically feasible mechanism. In Section~4, the
pressure angle, a key performance index of cam mechanisms, is
studied in order to choose the design parameters that give the
best pressure-angle distribution, a compromise being done with the
accuracy of the mechanism.
\section{Synthesis of the Planar Cam Mechanism}
Let the $x$-$y$ frame be fixed to the machine and the $u$-$v$
frame be attached to the cam, as depicted in Fig.~\ref{fig003}.
$O_{1}$ is the origin of both frames, while $O_{2}$ is the center
of the roller and $C$ is the contact point between cam and roller.
 \begin{figure}[htb]
 {\vspace{-0.5cm}}
 \begin{minipage}[b]{4.4cm}
 \begin{center}
   \psfrag{f}{$\bf f$}
   \psfrag{p}{$p$}    \psfrag{e}{$e$}    \psfrag{p}{$p$}    \psfrag{s}{$s$}
   \psfrag{d}{$d$}    \psfrag{x}{$x$}   \psfrag{y}{$y$}    \psfrag{P}{$P$}
   \psfrag{C}{$C$}    \psfrag{mu}{$\mu$}
   \psfrag{u}{$u$}       \psfrag{v}{$v$}
   \psfrag{delta}{$\delta$}
   \psfrag{b2}{$b_2$}   \psfrag{b3}{$b_3$}
   \psfrag{a4}{$a_4$}   \psfrag{psi}{$\psi$}
   \psfrag{theta}{$\theta$}
   \psfrag{O1}{$0_1$}   \psfrag{O2}{$0_2$}
   \centerline{\epsfig{file = 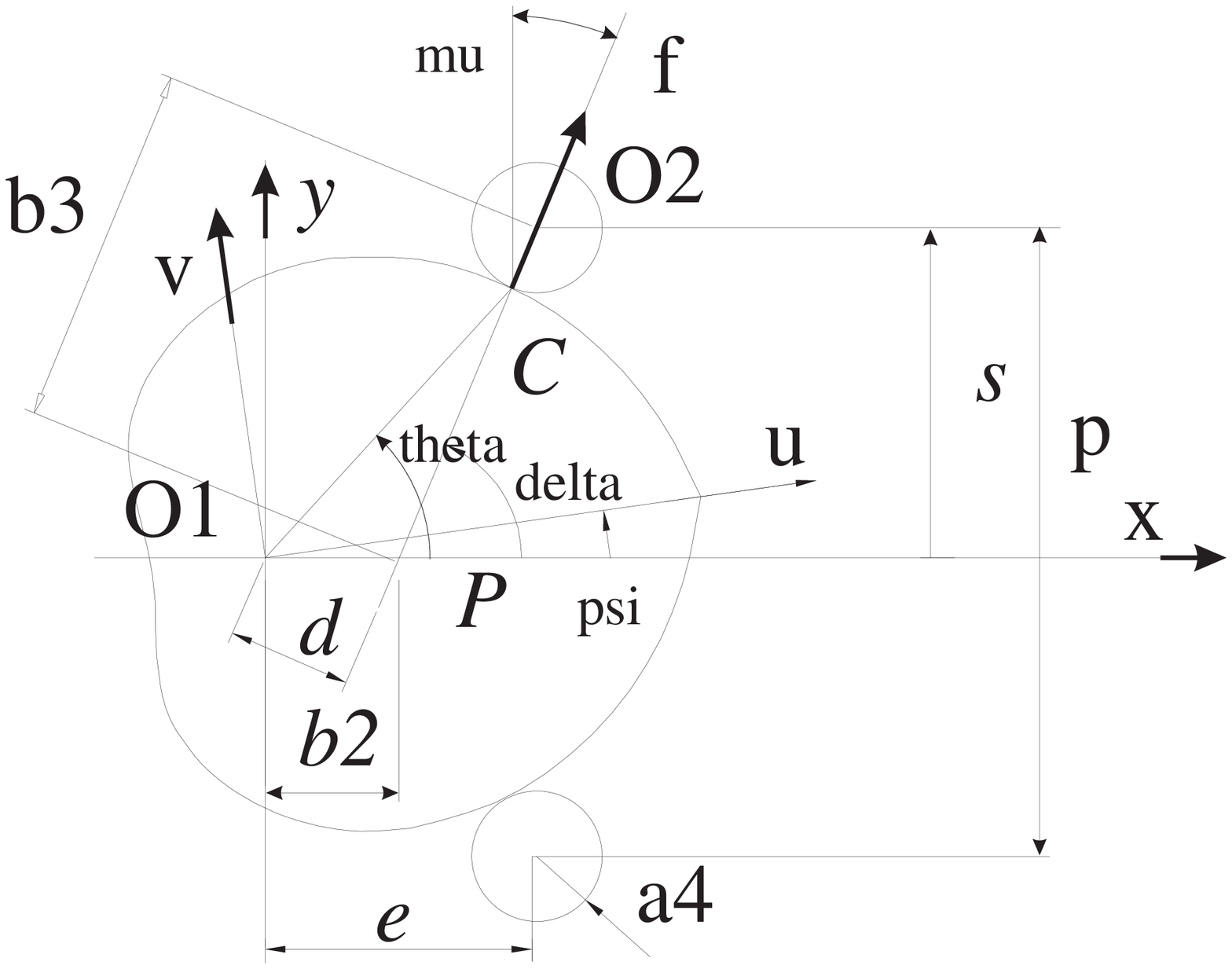,scale = 0.21}}
  \caption{Parameterization of \goodbreak Slide-O-Cam}
  \label{fig003}
 \end{center}
 \end{minipage}
 \begin{minipage}[b]{4.4cm}
 \begin{center}
   \psfrag{O1}{$0_1$}
   \psfrag{p}{$p$}    \psfrag{x}{$x$}    \psfrag{y}{$y$}
   \psfrag{u}{$u$}    \psfrag{v}{$v$}    \psfrag{x}{$x$}
   \psfrag{s(0)}{$s$(0)}
   \centerline{\epsfig{file = 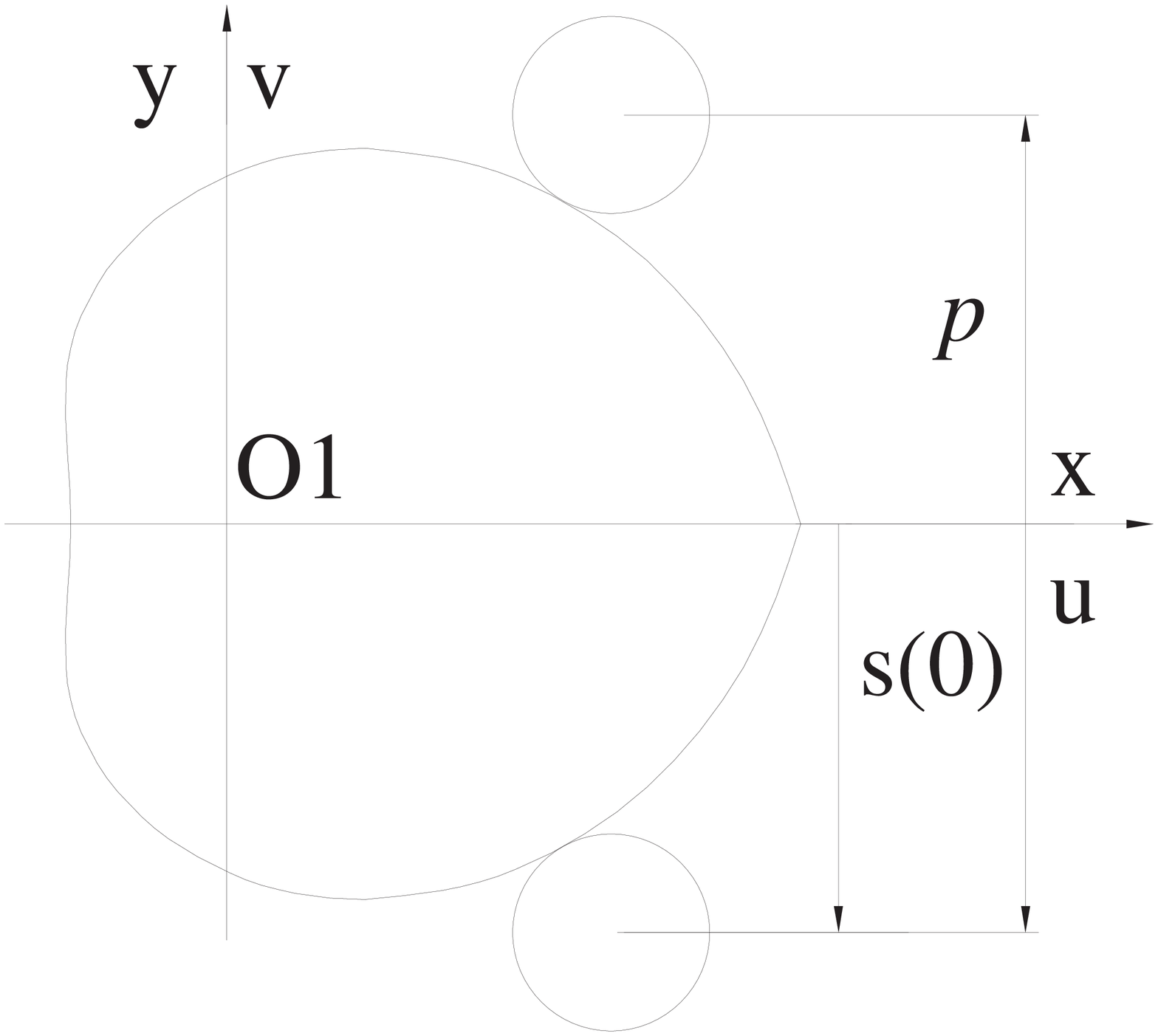,scale = 0.15}}
   \caption{Initial configuration of the mechanism} \label{fig004}
 \end{center}
 \end{minipage}
 {\vspace{-1cm}}
 \end{figure}
The geometric parameters defining the cam mechanism are
illustrated in the same figure. The notation of this figure is
based on the general notation introduced in
\cite{Gonzalez-Palacios:1993}, namely, (i) $p$: the pitch, {\it
i.e.}, the distance between the center of two rollers on the same
side of the follower; (ii) $e$: distance between the axis of the
cam and the line of centers of the rollers; (iii) $a_{4}$: radius
of the roller bearing, {\it i.e.}, the radius of the roller; (iv)
$\psi$: angle of rotation of the cam, the input of the mechanism;
(v) $s$: position of the center of the roller, {\it i.e}, the
displacement of the follower,  the output of the mechanism; (vi)
$\mu$: pressure angle; (vii) {\bf f}: force transmitted from the
cam to the roller. In this paper, $p$ is set to 50~mm, in order to
meet the Orthoglide specifications.

The above parameters as well as the contact surface on the cam,
are determined by the geometric relations dictated by the
Aronhold-Kennedy Theorem in the plane \cite{Waldron:1999}. When
the cam makes a complete turn ($\Delta \psi=2\pi$), the
displacement of the roller is equal to $p$, the distance between
two rollers on the same side of the roller-carrying slider
($\Delta s =p$). Furthermore, if we consider the initial
configuration of the roller as depicted in Fig.~\ref{fig004}, the
roller is on the lower side of the $x$-axis for $\psi=0$, so that
we have $s(0)=-p/2$. Hence, the input-output function $s$ is
 \begin{equation}
  s(\psi)=\frac{p}{2\pi}\psi-\frac{p}{2} \label{eq01}
 \end{equation}
The expression for the first and second derivatives of $s(\psi)$
with respect to $\psi$ will be needed:
 \begin{eqnarray}
 s'(\psi)=p/(2\pi) \quad {\rm and} \quad s''(\psi)=0
 \label{eq02}
 \end{eqnarray}
The cam profile is determined by the displacement of the contact
point $C$ around the cam. The Cartesian coordinates of this point
in the $u$-$v$ frame take the form \cite{Gonzalez-Palacios:1993}
 \begin{subequations}
 \begin{eqnarray}
 u_{c}(\psi) &=& ~~b_{2} \cos \psi+(b_{3}-a_{4})\cos(\delta-\psi) \\
 v_{c}(\psi) &=& -b_{2} \sin \psi + (b_{3}-a_{4})\sin(\delta-\psi)
 \end{eqnarray}
 \label{eq04}
 \end{subequations}
 with coefficients $b_{2}$, $b_{3}$ and $\delta$ given by
 \begin{subequations}
 \begin{eqnarray}
 b_{2} &=& -s'(\psi) \sin \alpha_{1} \\
 b_{3} &=& \sqrt{(e+s'(\psi)\sin \alpha_{1})^{2}+(s(\psi)
 \sin\alpha_{1})^{2}} \\
 \delta &=& \arctan \left( \frac{-s(\psi) \sin\alpha_{1}}{e+s'(\psi) \sin \alpha_{1}} \right)
 \end{eqnarray}
 \label{eq05}
 \end{subequations}
$\!\!\!$where $\alpha_{1}$ is the directed angle between the axis
of the cam and the translating direction of the follower;
$\alpha_{1}$ is positive in the ccw direction. Considering the
orientation adopted for the input angle $\psi$ and for the output
$s$, as depicted in Fig.~\ref{fig003}, we have
 \begin{equation}
 \alpha_{1}=-\pi /2
 \label{eq06}
 \end{equation}

We now introduce the nondimensional design parameter $\eta$, which
will be extensively used:
\begin{equation}
\eta= e/p \label{eq07}
\end{equation}

Thus, from Eqs.~(\ref{eq01}), (\ref{eq02}), (\ref{eq05}a-c),
(\ref{eq06}) and (\ref{eq07}), we compute the expressions for
coefficients $b_{2}$, $b_{3}$ and $\delta$ as
 \begin{subequations}
 \begin{eqnarray}
 b_{2} &=& \frac{p}{2\pi} \\
 b_{3} &=& \frac{p}{2\pi}\sqrt{(2\pi \eta -1)^{2}+(\psi-\pi)^{2}} \\
 \delta &=& \arctan\left(\frac{\psi-\pi}{2\pi \eta -1} \right)
 \end{eqnarray}
 \label{eq08}
 \end{subequations}
\newline whence a first constraint on $\eta$, $\eta \neq 1/(2\pi)$, is derived.
An \textit{extended angle} $\Delta$ is introduced \cite{Lee:2001},
so that the cam profile closes. Angle $\Delta$ is obtained as the
root of the equation $v_{c}(\psi)=0$.
%
In the case of Slide-O-Cam, $\Delta$ is negative, as shown in
Fig.~\ref{fig006}. Consequently, the cam profile closes within
$\Delta \leq \psi \leq 2\pi-\Delta$.
 \begin{figure}[!h]
 {\vspace{-0.5cm}}
 \begin{center}
     \psfrag{O1}{$O_1$}
     \psfrag{x}{$x$}
     \psfrag{y}{$y$}
     \psfrag{u}{$u$}
     \psfrag{v}{$v$}
     \psfrag{C}{$C$}
     \psfrag{Psi}{$\psi$}
     \subfigure[$\psi=\Delta$]{\epsfig{file =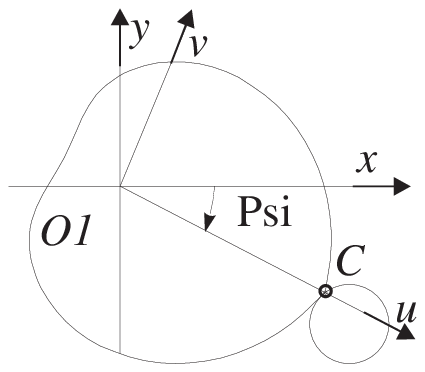,scale = 0.5}}
     \psfrag{x, u}{$x, u$}
     \psfrag{y, v}{$y, v$}
     \psfrag{delta}{$\Delta$}
     \subfigure[$\psi=0$]{\epsfig{file =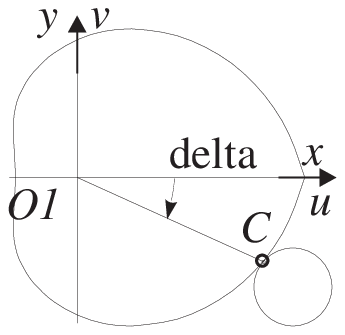,scale = 0.5}}
     \psfrag{-u}{-$u$}
     \psfrag{-v}{-$v$}
     \subfigure[$\psi=\pi$]{\epsfig{file =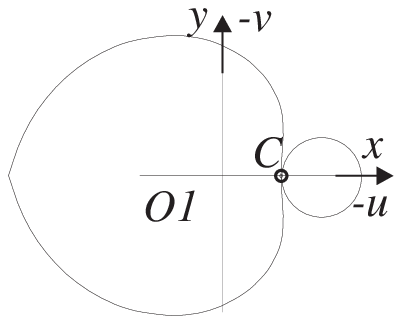,scale = 0.5}}
     \psfrag{u}{$u$}
     \psfrag{v}{$v$}
     \subfigure[$\psi=2\pi-\Delta$]{~~~\epsfig{file =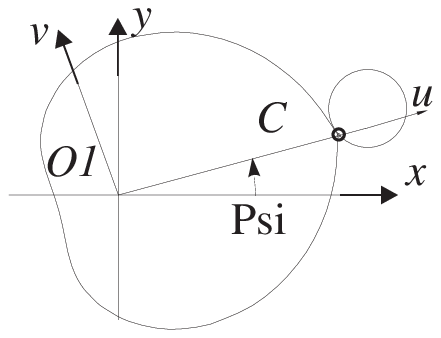,scale = 0.5}}
 \end{center}
 {\vspace{-0.5cm}}
 \caption{Extended angle $\Delta$}
 \label{fig006}
 {\vspace{-1cm}}
 \end{figure}
\subsection{Pitch-Curve Determination}
The pitch curve is the trajectory of the center $O_{2}$ of the
roller, distinct from the trajectory of the contact point $C$,
which produces the cam profile. The Cartesian coordinates of point
$O_{2}$ in the $x$-$y$ frame are $(e,s)$, as depicted in
Fig.~\ref{fig003}. Hence, the Cartesian coordinates of the
pitch-curve in the $u$-$v$ frame are
 \begin{subequations}
 \begin{eqnarray}
    u_{p}(\psi) & = & ~~e \cos \psi + s(\psi)\sin \psi
    \label{eq010}\\
    v_{p}(\psi) & = & -e  \sin \psi + s(\psi)\cos \psi
  \end{eqnarray}
 \end{subequations}
\subsection{Geometric Constraints on the Mechanism}
In order to lead to a feasible mechanism, the radius $a_{4}$ of
the roller must satisfy two conditions, as shown in
Fig.~\ref{fig010}a:

\noindent$\bullet$ Two consecutive rollers on the same side of the
roller-carrying slider must not be in contact. Since $p$ is the
distance between the center of two consecutive rollers, we have
the constraint $2a_{4} < p$. Hence the first condition on $a_{4}$:
     \begin{equation}
     \label{eq011}
     a_{4} / p< 1/2
     \end{equation}
$\bullet$ The radius $b$ of the shaft on which the cams are
mounted must be taken into consideration. Hence, we have the
constraint $a_{4}+b \leq e$, the second constraint on $a_{4}$ in
terms of the parameter $\eta$ thus being
     \begin{equation}
     \label{eq012}
     a_{4} / p \leq \eta -b / p
     \end{equation}
\begin{figure}[!ht]
 {\vspace{-0.5cm}}
\center
  \psfrag{O1}{$O_1$}
  \psfrag{x}{$x$}
  \psfrag{y}{$y$}
  \psfrag{p}{$p$}
  \psfrag{e}{$e$}
  \psfrag{b}{$b$}
  \psfrag{C}{$C$}
  \psfrag{a4}{$a_4$}
  \subfigure[]{\epsfig{file =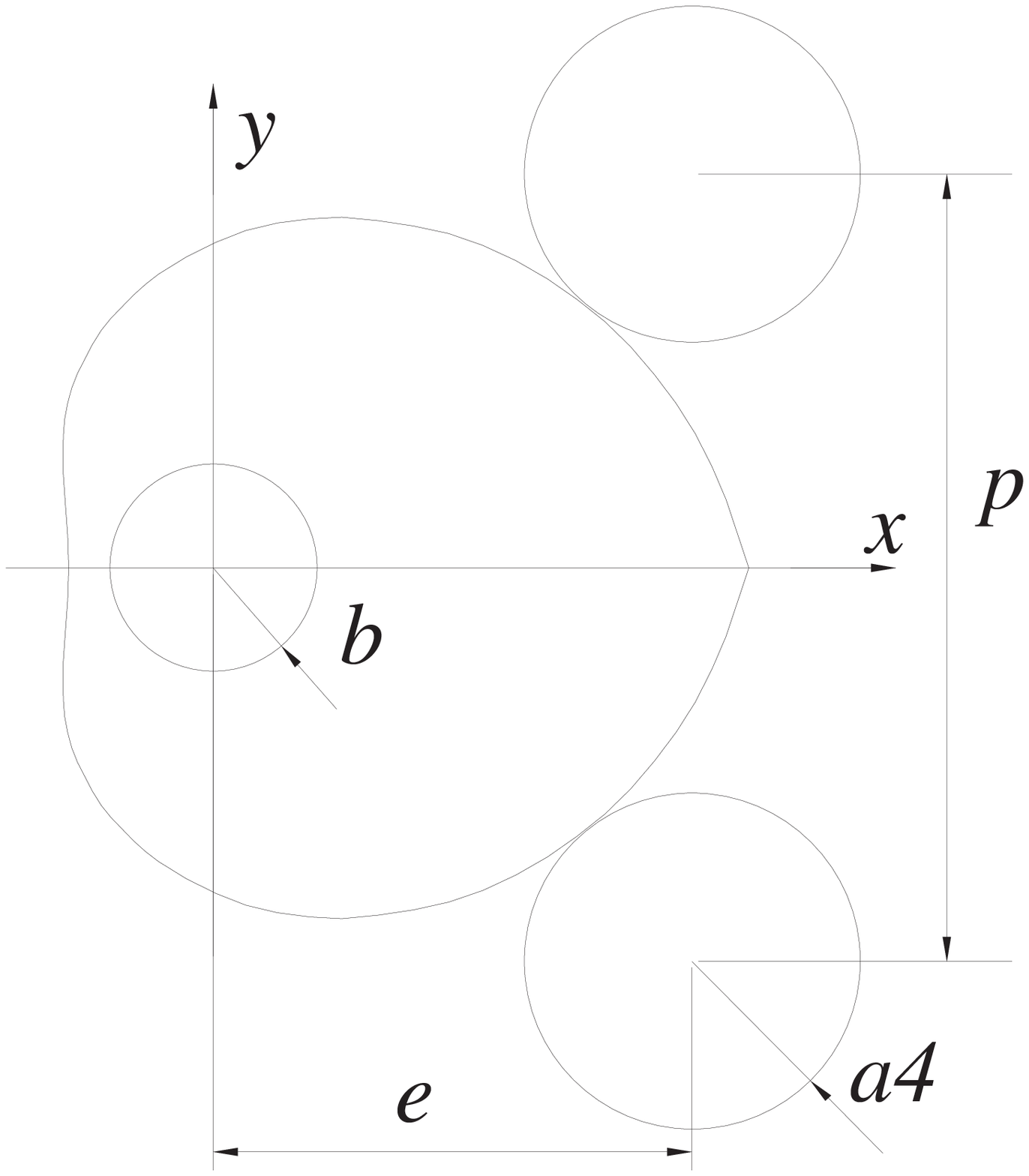,scale = 0.14}~~~~~~~}
  \subfigure[]{~~~~~~~~\epsfig{file =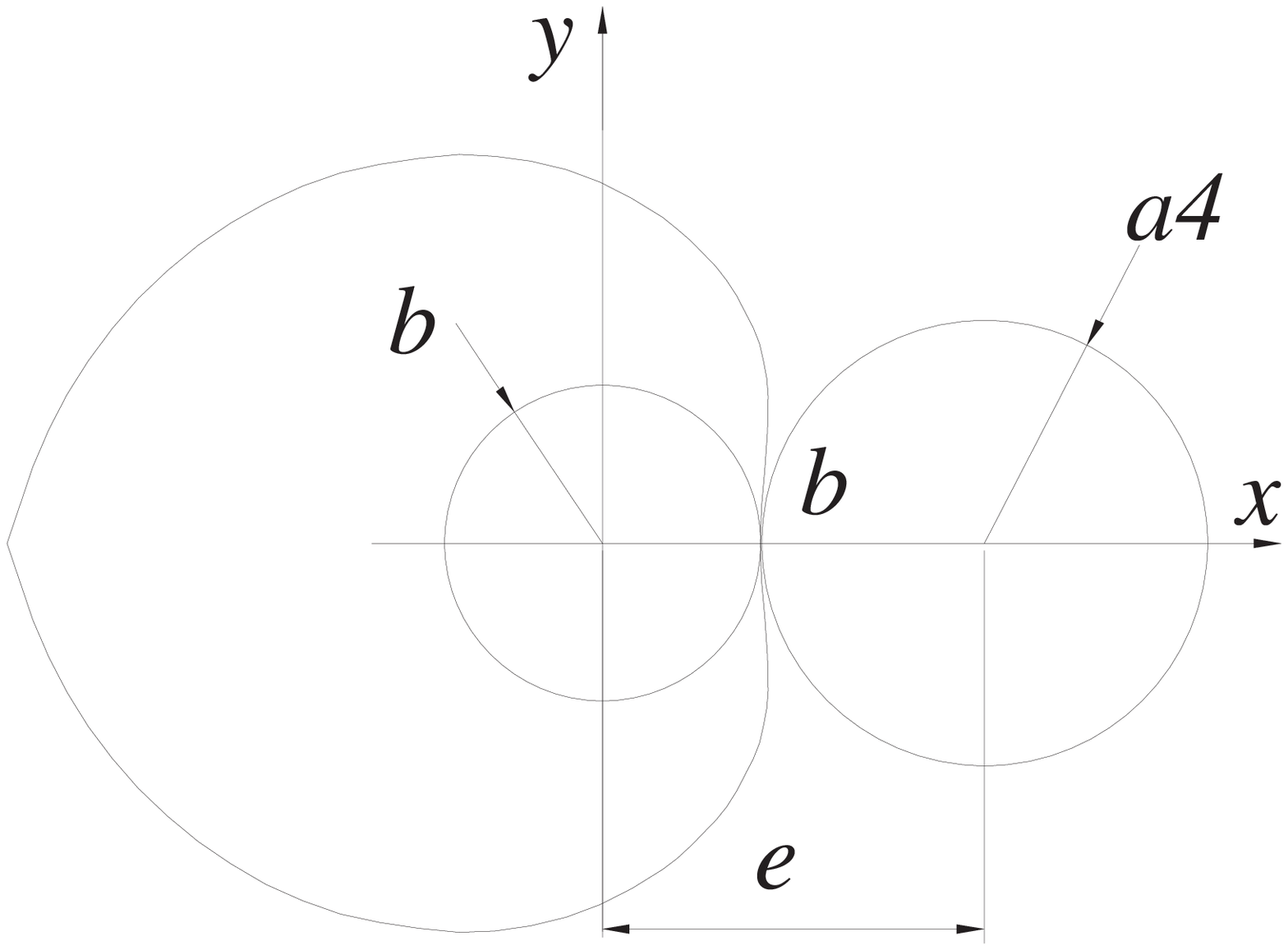,scale = 0.14}}
 {\vspace{-0.5cm}}
  \caption{Constraints on the radius of the roller} \label{fig010}
 {\vspace{-0.5cm}}
\end{figure}
Considering the initial configuration of the roller, as depicted
in Fig.~\ref{fig004}, the $v$-component of the Cartesian
coordinate of the contact point $C$ is negative in this
configuration, {\it i.e.}, $ v_{c}(0) \leq 0 $. Considering the
expression for $v_{c}(\psi)$ and for parameters $b_{3}$ and
$\delta$ given in Eqs.~(\ref{eq04}b), (\ref{eq08}b \& c),
respectively, the above relation leads to the condition:
 \beqa
  \left( \frac{p}{2\pi a_4}\sqrt{(2\pi \eta-1)^{2}+(-\pi)^{2}} - 1 \right)
  \sin \left[\arctan\left(\frac{-\pi}{2\pi \eta -1}\right) \right]
  \leq 0 \nonumber
  \eeqa
Further, we define $A$ and $B$ as:
 \beqa
 A=\frac{p}{2\pi a_4}\sqrt{(2\pi \eta-1)^{2}+\pi^{2}} - 1
 ~\mbox{and}~
 B=\sin\left[\arctan\left(\frac{-\pi}{2\pi \eta -1}\right) \right]
 \nonumber
 \eeqa
Since $(2\pi \eta-1)^{2} >0$, we have,
 \beqa
    \sqrt{(2\pi \eta-1)^{2}+\pi^{2}} > \pi
 \eeqa
Hence, $ A > p/(2a_4) - 1$. Furthermore, from the constraint on
$a_{4}$, stated in Eq.~(\ref{eq011}), we have $p/(2a_4)-1
> 0$, whence $A>0$. Consequently, the constraint $v_{c}(0) \leq 0$
leads to the constraint $B \leq 0$.
We rewrite the expression for $B$, by using the trigonometric
relation,
 \beqa
    B \leq 0 \Longleftrightarrow \frac{-\pi}{(2\pi \eta-1) \sqrt{1+\pi^{2}/(2\pi \eta-1)^{2}}} \leq 0
    \nonumber
 \eeqa
which holds only if $2\pi \eta-1 > 0$. Finally, the constraint
$v_{c}(0) \leq 0$ leads to a constraint on $\eta$:
\begin{equation}
\label{eq013} \eta> 1/(2\pi)
\end{equation}
\subsection{Pressure Angle}
The pressure angle is defined as the angle between the common
normal at the cam-roller contact point $C$ and the velocity of the
follower \cite{Angeles:1991}, as depicted in Fig.~\ref{fig003},
where the presure angle is denoted by $\mu$. This angle plays an
important role in cam design. The smaller $|\mu|$, the better the
force transmission. In the case of high-speed operations, {\it
i.e.}, angular velocities of cams exceeding 50~rpm, the
recommended bounds of the pressure angle are within 30$^{\circ}$.
Nevertheless, as it is not always possible to have a pressure
angle that remains below 30$^{\circ}$, we adopt the
\textit{service factor}, which is the percentage of the working
cycle with a pressure angle within 30$^{\circ}$ \cite{Lee:2001}.
The service angle will be useful to take into consideration these
notions in the ensuing discussion, when optimizing the mechanism.

For the case at hand, the expression for the pressure angle $\mu$
is given in \cite{Angeles:1991} as
 \[
 \mu=\arctan \left( \frac{s'(\psi)-e}{s(\psi)} \right)
 \]
Considering the expressions for $s$ and $s'$, and using the
parameter $\eta$ given in Eqs.~(\ref{eq01}), (\ref{eq02}a) and
(\ref{eq07}), respectively, the expression for the pressure angle
becomes
 \begin{equation}
 \mu=\arctan \left( \frac{1-2\pi \eta}{\psi - \pi} \right)
 \label{eq0111}
 \end{equation}
We are only interested in the value of the pressure angle with the
cam driving the roller, which happens with
 \begin{equation}
   \label{eq01111} \pi \leq \psi \leq 2\pi-\Delta
 \end{equation}
Indeed, if we start the motion in the initial configuration
depicted in Fig.~\ref{fig006}b, with the cam rotating in the ccw
direction, the cam begins to drive the roller only when
$\psi=\pi$; and the cam can drive the follower until contact is
lost, {\it i.e.}, when $\psi=2\pi-\Delta$, as shown in
Figs.~\ref{fig006}c \& d.

Nevertheless, as shown in Fig.~\ref{fig008}, the conjugate cam can
also drive the follower when $0 \leq \psi \leq \pi-\Delta$; there
is therefore a common interval, for $\pi \leq \psi \leq
\pi-\Delta$, during which two cams can drive the follower. In this
interval, the conjugate cam can drive a roller with lower absolute
values of the pressure angle. We assume that, when the two cams
can drive the rollers, the cam with the lower absolute value of
pressure angle effectively drives the follower. Consequently, we
are only interested in the value of the pressure angle in the
interval,
 \begin{equation}
 \label{eq0080} \pi-\Delta \leq \psi \leq 2\pi-\Delta
 \end{equation}

We study here the influence of parameters $\eta$ and $a_{4}$ on
the values of the pressure angle while the cam drives the roller,
{\it i.e.}, with $\pi-\Delta \leq \psi \leq 2\pi-\Delta$, as
explained above.

$\bullet$ {\it Influence of parameter $\eta$}: Figure~\ref{fig30}
shows the influence of the parameter $\eta$ on the pressure angle,
with $a_{4}$ and $p$ being fixed. From these plots we have one
result: {\it The lower $\eta$, the lower $|\mu|$.}

$\bullet$ {\it Influence of the radius of the roller $a_{4}$:}
$a_{4}$ does not appear in the expression for the pressure angle,
but it influences the value of the extended angle $\Delta$, and
hence, the plot boundaries of the pressure angle, as shown in
Fig.~\ref{fig008}.

By computing the value of the extended angle $\Delta$ for several
values of $a_{4}$, we can say that the higher $a_{4}$, the lower
$|\Delta|$. Consequently, since the boundaries to plot the
pressure angle are $\pi-\Delta$ and $2\pi-\Delta$, we can say that
when we increase $a_{4}$, $-\Delta$ decreases and the boundaries
are translated toward the left, {\it i.e.}, toward higher absolute
values of the pressure angle.
\begin{figure}[!hb]
 {\vspace{-0.5cm}}
 \begin{center}
 \begin{minipage}[b]{4.4cm}
 \psfrag{mu}{$\mu$}
 \psfrag{psi}{$\psi$}
 {\scriptsize
 \psfrag{pi}{$\pi$}
 \psfrag{pi-Delta}{$\pi-\Delta$}
 \psfrag{2pi-Delta}{$2\pi-\Delta$}
 \psfrag{-6}{-6}
 \psfrag{-2}{-2}
 \psfrag{2}{2}
 \psfrag{6}{6}
 \psfrag{10}{10}
 \psfrag{100}{100}
 \psfrag{50}{50}
 \psfrag{-50}{-50}
 \psfig{file= 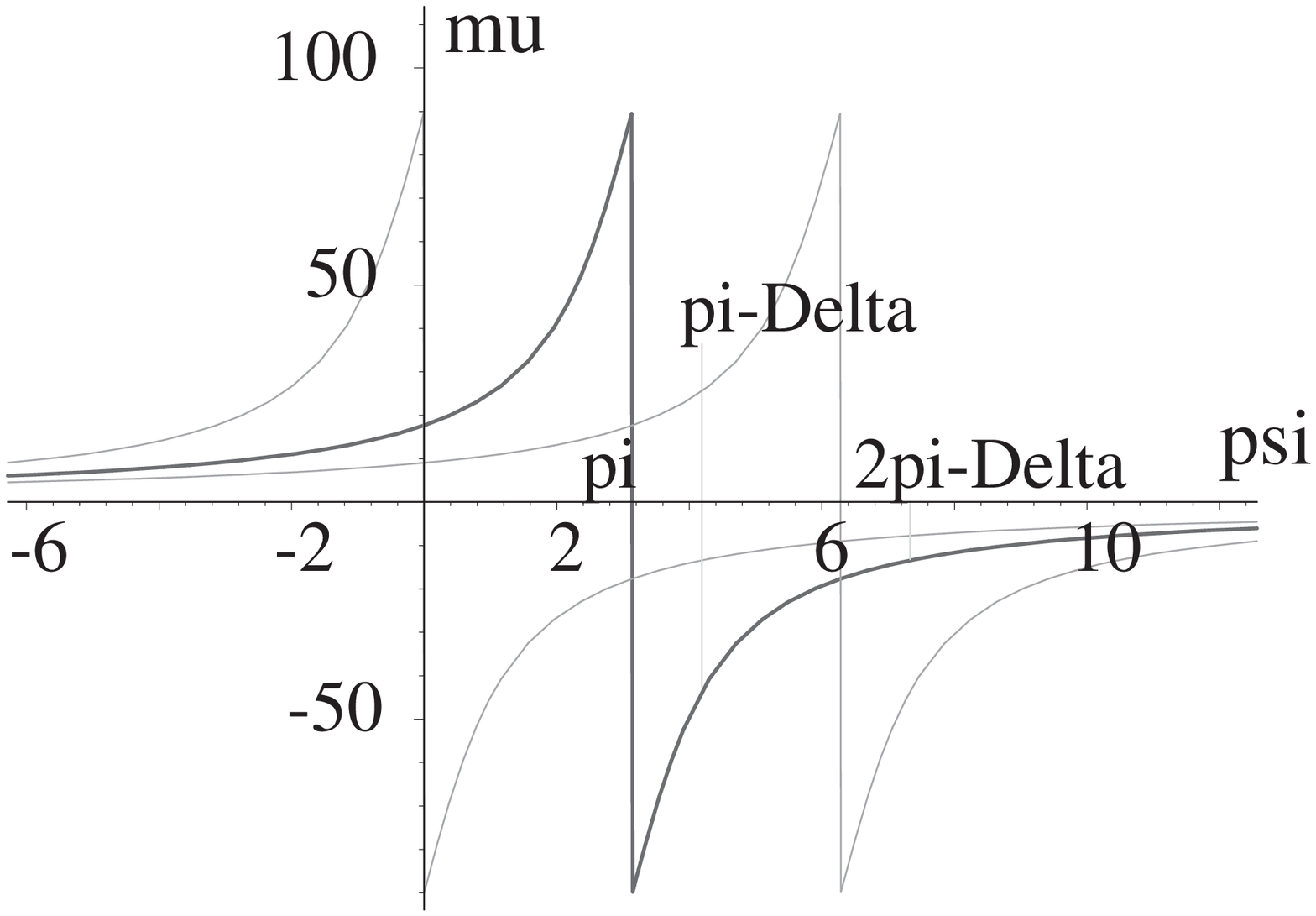, scale=
 0.2}}
 \caption{Pressure angle distribution}  \label{fig008}
 \end{minipage}
 \begin{minipage}[b]{4.4cm}
 \psfrag{mu}{$\mu$}
 \psfrag{psi}{$\psi$}
 \psfrag{eta}{$\eta$}
 {\scriptsize
 \psfrag{8}{8}  \psfrag{6}{6}  \psfrag{4}{4}  \psfrag{2}{2}
 \psfrag{-20}{-20}  \psfrag{-40}{-40}  \psfrag{-60}{-60}  \psfrag{-80}{-80}
 \psfig{file= 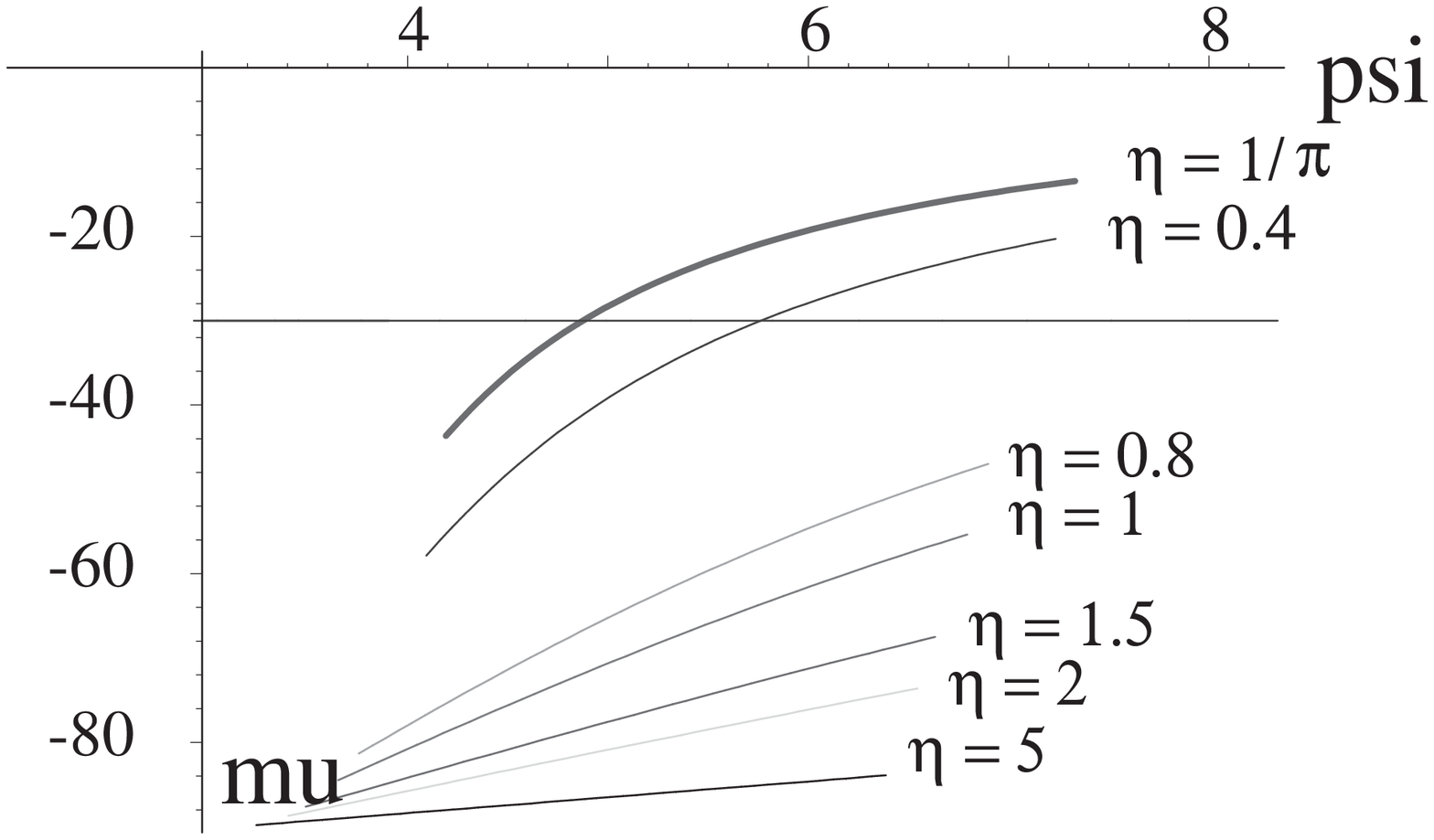, scale=
 0.2}}
 \caption{Influence of parameter $\eta$ on the pressure angle
 $\mu$ (in degree), with $p=50$~mm and $a_{4}=10$~mm} \label{fig30}
 \end{minipage}
 \end{center}
 {\vspace{-1.5cm}}
 \end{figure}
\section{Convexity of the Cam Profile and Undercutting}
In order to enhance machining accuracy, we need the cam profile to
be fully convex. In this section we establish conditions on the
design parameters $\eta$ and $a_{4}$ in order to have a fully
convex cam profile. So, we study the sign of the curvature of the
cam profile via that of the pitch curve. Furthermore, for cam
design in roller-follower mechanisms, we should also consider
\textit{undercutting}. Undercutting occurs when the radius of the
roller is greater than or equal to the minimum absolute value of
the radius of curvature of the pitch curve. Upon avoiding
undercutting, the  sign of the curvature of the pitch curve is
identical to that of the cam profile.
\subsection{Curvature of the Cam Profile}
The curvature of any planar parametric curve, in terms of the
Cartesian coordinates $u$ and $v$, and parameterized with any
parameter $\psi$, is given by \cite{Angeles:1991}:
 \begin{equation}
 \label{eq1}
 \kappa=\frac{v'(\psi)u''(\psi)-u'(\psi)v''(\psi)}{[u'(\psi)^{2}+v'(\psi)^{2}]^{3/2}}
 \end{equation}
The sign of $\kappa$ in Eq.~(\ref{eq1}) tells whether the curve is
convex or concave at a point: a positive $\kappa$ implies a
convexity, while a negative $\kappa$ implies a concavity at that
point. To obtain the curvature of the cam profile for a given
roller-follower, we use the Cartesian coordinates of the pitch
curve, since obtaining its first and second derivatives leads to
simpler expressions as compared with those associated with the cam
profile itself. Then, the curvature of the cam profile is derived
by a simple geometric relationship between the curvatures of the
pitch curve and of the cam profile.

The Cartesian coordinates of the pitch curve were recalled in
Eqs.~(\ref{eq010} \& b), while Eqs.~(\ref{eq02}a \& b) give their
first and second derivatives
 \beqa
 s(\psi)= p/(2\pi)\psi-p/2~,~~~
 s'(\psi)  =  p / (2\pi)~,~~~
 s''(\psi)  =  0 \nonumber
 \eeqa
With the above-mentioned expressions, we can compute the first and
second derivatives of the Cartesian coordinates of the pitch curve
with respect to the angle of rotation of the cam, $\psi$:
 \begin{subequations}
 \begin{eqnarray}
 u'_{p}(\psi) & = & ~~~(~~s'(\psi)-e)\sin \psi + s(\psi)\cos \psi \label{eq2} \\
 v'_{p}(\psi) & = & ~~~(~~s'(\psi)-e)\cos \psi - s(\psi)\sin \psi  \\
 u''_{p}(\psi) & = & ~~~(2s'(\psi)-e)\cos \psi - s(\psi)\sin \psi  \\
 v''_{p}(\psi) & = & -(2s'(\psi)-e)\sin \psi - s(\psi)\cos \psi
 \end{eqnarray}
 \end{subequations}
By substituting $\eta$, $\eta=e/p$, along with Eqs.~(\ref{eq2}-d),
into Eq.~(\ref{eq1}), the curvature $\kappa_{p}$ of the pitch
curve is obtained as
 \begin{equation}
 \label{eq6}
 \kappa_{p}=\frac{2\pi}{p} \frac{[(\psi-\pi)^{2}+2(2\pi
\eta-1)(\pi \eta-1)]}{[(\psi-\pi)^{2}+(2\pi \eta-1)^{2}]^{3/2}}
 \end{equation}
provided that the denominator never vanishes for any value of
$\psi$, {\it i.e.}, provided that
 \begin{equation}
 \label{eq7}
 \eta \neq 1/(2\pi)
 \end{equation}
Let $\rho_{c}$ and $\rho_{p}$ be the radii of curvature of both
the cam profile and the pitch curve, respectively, and
$\kappa_{c}$ the curvature of the cam profile. Since the curvature
is the reciprocal of the radius of curvature, we have $\rho_{c} =
1/\kappa_{c}$ and $\rho_{p} = 1/\kappa_{p}$. Furthermore, due to
the definition of the pitch curve, it is apparent that
 \begin{equation}
 \label{eq8} \rho_{p} = \rho_{c} + a_{4}
 \end{equation}
Writing Eq.~(\ref{eq8}) in terms of $\kappa_{c}$ and $\kappa_{p}$,
we obtain the curvature of the cam profile as
 \begin{equation}
 \label{eq9} \kappa_{c}=\frac{\kappa_{p}}{1-a_{4} \kappa_{p}}
 \end{equation}
with $\kappa_{p}$ given in Eq.~(\ref{eq6}). As we saw previously,
we want the cam profile to be fully convex, which happens if the
pitch curve is fully convex too. We thus find first the convexity
condition of the pitch curve.
\subsection{Convexity Condition of the Pitch Curve}
Considering the expression for $\kappa_{p}$ in Eq.~(\ref{eq6}), we
have, for every value of $\psi$, $\kappa_{p} \geq 0$ if  $(2\pi
\eta-1)(\pi \eta-1) \geq 0$ and $\eta \neq 1 / (2\pi)$, whence the
condition on $\eta$:
 \begin{equation}
 \label{eq10}
 \kappa_{p} \geq 0
 ~~~{\rm if}~~~
 \eta \in [0, 1/(2\pi)[ ~~\cup~~ [1/\pi, + \infty[
 \end{equation}
Figure~\ref{fig100} shows pitch curve profiles and their
curvatures for two values of $\eta$.
\begin{figure}[!ht]
 {\vspace{-0.5cm}}
 \begin{minipage}[b]{4.2cm}
 {\scriptsize
  \psfrag{-20}{-$20$}
  \psfrag{-10}{-$10$}
  \psfrag{10}{$10$}
  \psfrag{20}{$20$}
  \psfrag{30}{$30$}
  \subfigure[$\eta=0.2$ ($\eta \in \rbrack \frac{1}{2\pi},\frac{1}{\pi} \lbrack $)]{~~~~~~~~~~~~~~~~~\psfig{file=
  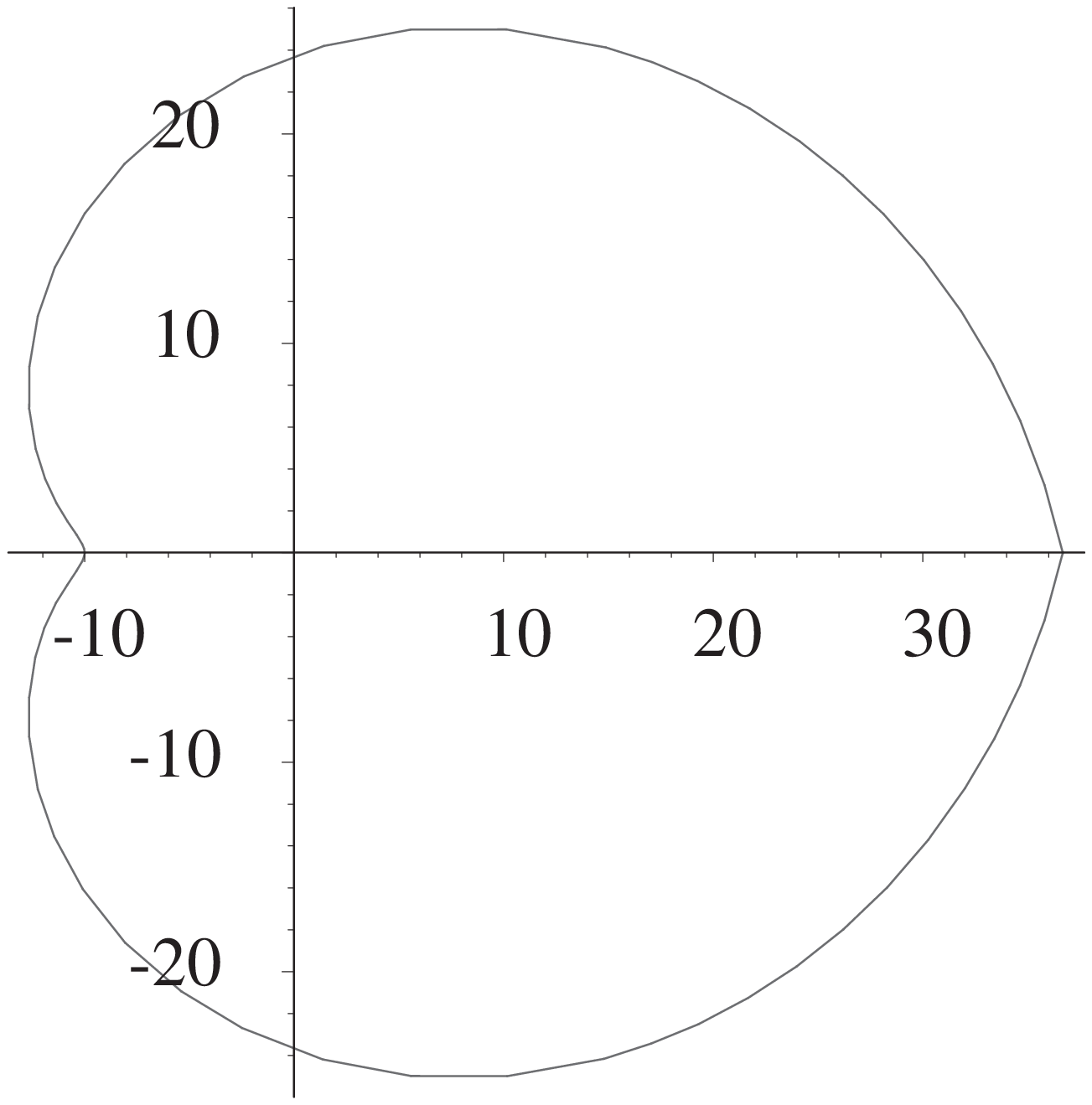,scale=0.17}~~~~~~~~~~~~~~~~~~~~~}}
 \end{minipage}
 \begin{minipage}[b]{4.2cm}
 {\scriptsize
  \psfrag{-20}{-$20$}
  \psfrag{-40}{-$40$}
  \psfrag{40}{$40$}
  \psfrag{20}{$20$}
  \psfrag{30}{$30$}
  \subfigure[$\eta=0.7$ ($\eta>1/\pi$)]{~~~~~~~~~\psfig{file= 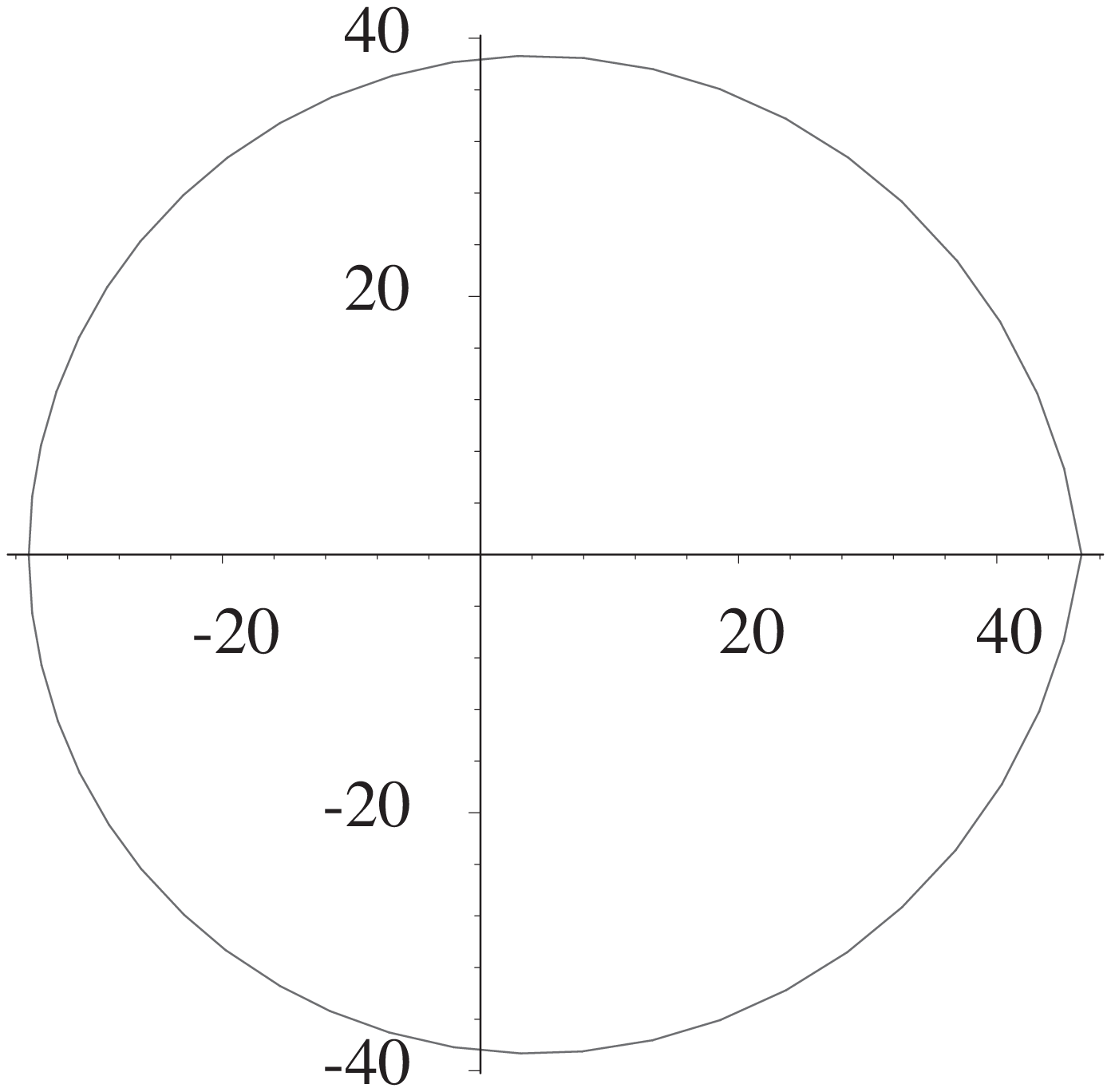,
  scale=0.17}~~~~~~~~~~~~~~~~}}
 \end{minipage}
 {\vspace{-0.5cm}}
 \begin{minipage}[b]{4.2cm}
 \begin{center}
  \psfrag{psi}{$\psi$}
  \psfrag{kappa_p}{$\kappa_p$}
  {\scriptsize
   \psfrag{0.2}{0.2}
   \psfrag{-0.2}{-0.2}
   \psfrag{-0.4}{-0.4}
   \psfrag{-0.6}{-0.6}
   \psfrag{-0.8}{-0.8}
   \psfrag{-1.0}{-1.0}
   \psfrag{-1.2}{-1.2}
   \psfrag{-1.4}{-1.4}
   \psfrag{2}{2}
   \psfrag{4}{4}
   \psfrag{6}{6}
  \subfigure[Pitch curve curvature with $\eta=0.2$ ($\eta \in \rbrack \frac{1}{2\pi},\frac{1}{\pi} \lbrack $)]{\psfig{file= 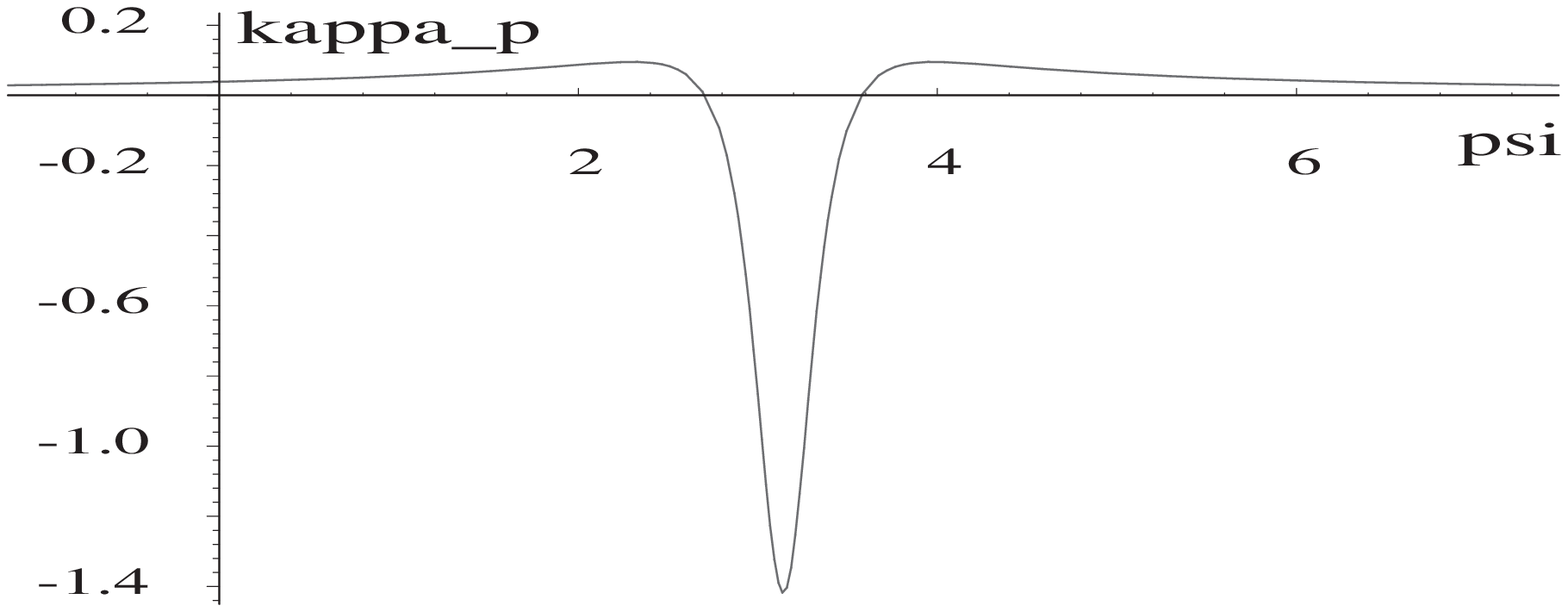, scale=
  0.2}}}
 \end{center}
 \end{minipage}
 \begin{minipage}[b]{4.2cm}
 \begin{center}
  \psfrag{psi}{$\psi$}
  \psfrag{kappa_p}{$\kappa_p$}
  {\scriptsize
   \psfrag{0.022}{0.022}
   \psfrag{0.023}{0.023}
   \psfrag{0.024}{0.024}
   \psfrag{0.025}{0.025}
   \psfrag{0.026}{0.026}
   \psfrag{0}{0}    \psfrag{1}{1}    \psfrag{2}{2}    \psfrag{3}{3}
   \psfrag{4}{4}    \psfrag{5}{5}    \psfrag{6}{6}    \psfrag{7}{7}
  \subfigure[Pitch curve curvature with $\eta=0.7$ ($\eta>1/\pi$)]{\psfig{file= 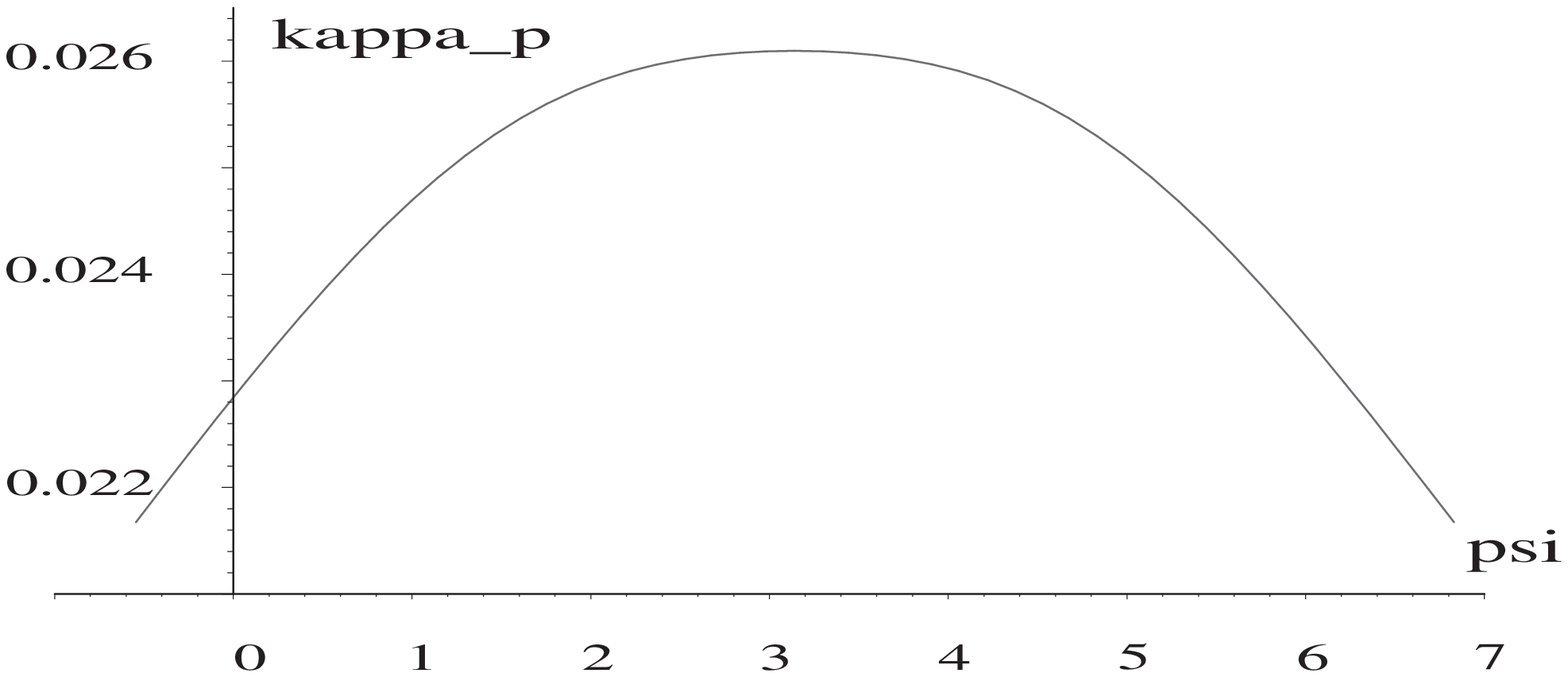,
  scale=0.2}}}
 \end{center}
 \end{minipage}
  \caption{Concave (a) and convex (b) pitch curve profiles and their corresponding curvatures with $p=50$~mm}
  \label{fig100}
 {\vspace{-0.5cm}}
\end{figure}
The condition on $\eta$ given in Eq.~(\ref{eq10}) must be combined
with the condition appearing in Eq.~(\ref{eq013}), $\eta
> 1/(2\pi)$; hence, the final \textbf{convexity condition of the
pitch curve} is:
 \begin{equation}
 \label{eq11}
 \eta \geq 1/\pi
 \end{equation}
\subsection{Undercutting Avoidance}
We assume in this subsection that the pitch curve is fully convex,
{\it i.e.}, $\kappa_{p} \geq 0$ and $\eta \geq 1/\pi$. In order to
avoid undercutting, {\it i.e.}, in order to have both the cam
profile and the pitch curve fully convex, we need $\kappa_{c}$ to
be positive. Considering the expression for the curvature of the
cam profile $\kappa_{c}$ of Eq.~(\ref{eq9}), the condition to
avoid undercutting is $1-a_{4} \kappa_{p} > 0$, whence the
condition on the radius of the follower $a_{4}$ is
 \[
 a_{4} < \frac{1}{\kappa_{p}(\psi)} ~~~~~ \forall \psi \in \mathbb{R}
 \]
Since $\kappa_{p}$ is positive, this condition can be written as
 \begin{equation}
 \label{eq12}
 a_{4} < \frac{1}{\kappa_{p\rm max}}
 ~~~~~ \mbox{with}~~~
 \kappa_{p\rm max}= \max_{\psi \in \mathbb{R}} \kappa_{p}(\psi)
 \end{equation}

$\bullet$ {\it Expression for $\kappa_{p\rm max}$}: In order to
compute the expression for $\kappa_{p\rm max}$, we need the first
derivative $\kappa'_{p}$ of $\kappa_{p}$ with respect to $\psi$
and its roots.  With the condition $\eta \geq 1/\pi$, the
expression for $\kappa_{p}$ given in Eq.~(\ref{eq6}) is
differentiable for every value of $\psi$. Thus, we obtain
 \begin{equation}
 \kappa'_{p}=- \frac{2 \pi}{p} \frac{(\psi-\pi)(\psi^{2}
-2\pi\psi+\pi^{2}+4\eta^{2}\pi^{2}-10\eta\pi+4)}{[(\psi-\pi)^{2}+(2\pi
\eta-1)^{2}]^{5/2}} \nonumber
 \end{equation}
The roots of $\kappa'_{p}$ are, apparently, $\psi_{1}=\pi$ and the
roots $\psi_{2}$ and $\psi_{3}$ of the polynomial
 \[
 P(\psi)=\psi ^{2} -2\pi\psi+\pi^{2}+4\eta^{2}\pi^{2}-10\eta\pi+4
 \]
Let $\beta_{\psi}$ be the discriminant of the equation $P=0$, {\it
i.e.},
 \[
 \beta_{\psi}=-4\eta^{2}\pi^{2}+10\eta\pi-4
 \]
Therefore, the sign of $\beta_{\psi}$ and, consequently, the roots
$\psi_{2}$ and $\psi_{3}$, depend on the value of $\eta$. Let
$\beta_{\eta}$ be the discriminant of $\beta_{\psi}=0$, a
quadratic equation in $\eta$. Hence, $\beta_{\eta}=9\pi^{2}$,
which is positive. The two roots of $\beta_{\psi}$ are $1/2\pi$
and $2/\pi$. Thus,
 \begin{eqnarray*}
 \beta_{\psi}>0 ~~~&\mbox{if}&~~~ \eta \in [\frac{1}{\pi},\frac{2}{\pi}[
 \quad {\rm or} \quad
 \beta_{\psi}<0 ~~~\mbox{if}~~~ \eta \in ] \frac{2}{\pi},+\infty[ \\
 \beta_{\psi}=0 ~~~&\mbox{if}&~~~ \eta = \frac{2}{\pi}
 \end{eqnarray*}
We now study the roots of $\kappa'_{p}$ according to the value of
$\eta$.

\noindent$\bullet$ $\eta \in [ 1/\pi,2/\pi[$: $\beta_{\psi}>0$,
and the polynomial $P$ has two roots $\psi_{2}$ and $\psi_{3}$, so
that $\kappa'_{p}$ has three roots:
     \begin{subequations}
     \begin{eqnarray}
     \psi_{1} &=& \pi \\
     \psi_{2} &=& \pi + \sqrt{-4\eta^{2}\pi^{2}+10\eta\pi-4} \\
     \psi_{3} &=& \pi - \sqrt{-4\eta^{2}\pi^{2}+10\eta\pi-4}
     \end{eqnarray}
     \label{eq14}
     \end{subequations}
\noindent$\bullet$ $\eta \in ] 2/\pi,+\infty[$: $\beta_{\psi}<0$,
and the polynomial $P$ has no real roots, so that $\kappa'_{p}$
has only one root, $\psi_{1}=\pi$.

\noindent$\bullet$  $\eta=2/\pi$: $\beta_{\psi}=0$, and the
polynomial $P$ has one double root equal to $\pi$, so that
$\kappa'_{p}$ has one triple root $\psi_{1}=\pi$.

To decide whether these roots correspond to minima or maxima, we
need to know the sign of the second derivative $\kappa''_{p}$ of
$\kappa_{p}$ with respect to $\psi$, for the corresponding values
of $\psi$. If the second derivative is negative, the stationary
value is a maximum; if positive, a minimum. The expressions for
the second derivatives were computed with Maple, for the values of
$\psi$ given in Eqs.~(\ref{eq14}a-c):
\begin{eqnarray*}
 \kappa''_{p}(\psi_{1}) &=& \frac{4\pi (\eta\pi -2)}{p |2\eta\pi-1|^{3}(2\eta\pi-1)} \\
 \kappa''_{p}(\psi_{2})=\kappa''_{p}(\psi_{3}) &=& \frac{8\pi (\eta\pi -2)}{9p(2\eta\pi-1) \sqrt{6\eta\pi-3}}
\end{eqnarray*}

$\bullet$  If $\eta \in [ 1/\pi,2/\pi[$,
$\kappa''_{p}(\psi_{1})>0$ and
$\kappa''_{p}(\psi_{2})=\kappa''_{p}(\psi_{3})<0$, the curvature
of the pitch curve has one local minimum for $\psi_{1}$ and two
maxima, for $\psi_{2}$ and $\psi_{3}$. Hence, the value of
$\kappa_{p \rm max}$ is
   \begin{equation}
   \label{eq19}
   \kappa_{p \rm max1}=\kappa_{p}(\psi_{2})=\kappa_{p}(\psi_{3})=\frac{4\pi}{3p \sqrt{6\eta\pi-3}}
   \end{equation}
Figure~\ref{fig101}a shows a plot of the pitch-curve curvature
with $\eta=1/\pi$ and $p=50$~mm. Since $\eta$ is taken equal to
the convexity limit $1/\pi$, the curvature remains positive and
only vanishes for $\psi=\pi$.

$\bullet$ If $\eta \in ] 2/\pi,+\infty[$,
$\kappa''_{p}(\psi_{1})<0$, the curvature of the pitch curve has a
maximum for $\psi_{1}$. Hence, the value of $\kappa_{p \rm max}$
is
   \begin{equation}
   \label{eq20}
   \kappa_{p \rm max2}=\kappa_{p}(\psi_{1})=\frac{4\pi}{p}\frac{
   (2\eta^{2}\pi^{2}-3\eta\pi+1)}{(4\eta^{2}\pi^{2}-4\eta\pi+1)^{3/2}}
   \end{equation}
Figure~\ref{fig100}d shows a plot of the pitch-curve curvature
with $\eta=0.7$ ($\eta>2/\pi$) and $p=50$~mm.

$\bullet$ If $\eta=2/\pi$, $\kappa''_{p}(\psi_{1})=0$, we cannot
tell whether we are in the presence of a maximum or a minimum. We
solve this uncertainty graphically, by plotting the curvature of
the pitch curve for $\eta=2/\pi$ and $p=50$~mm.
Figure~\ref{fig101}b reveals that the curvature has a maximum for
$\psi_{1}$. The value of this maximum can be obtained by
substituting $\eta$ by $2/\pi$ into either $\kappa_{p \rm max1}$
or $\kappa_{p \rm max2}$, expressed in Eqs.~(\ref{eq19}) and
(\ref{eq20}), respectively.

\begin{figure}
 {\vspace{-0.4cm}}
 \begin{center}
 \begin{minipage}[b]{4.45cm}
  \begin{center}
   \psfrag{kappa_p}{$\kappa_p$}
   \psfrag{psi}{$\mu$ (rad)}
   {\scriptsize
   \psfrag{2}{2}
   \psfrag{4}{6}
   \psfrag{6}{6}
   \psfrag{0.02}{0.02}
   \psfrag{0.04}{0.04}
   \subfigure[$\eta=1/\pi$]{\psfig{file= 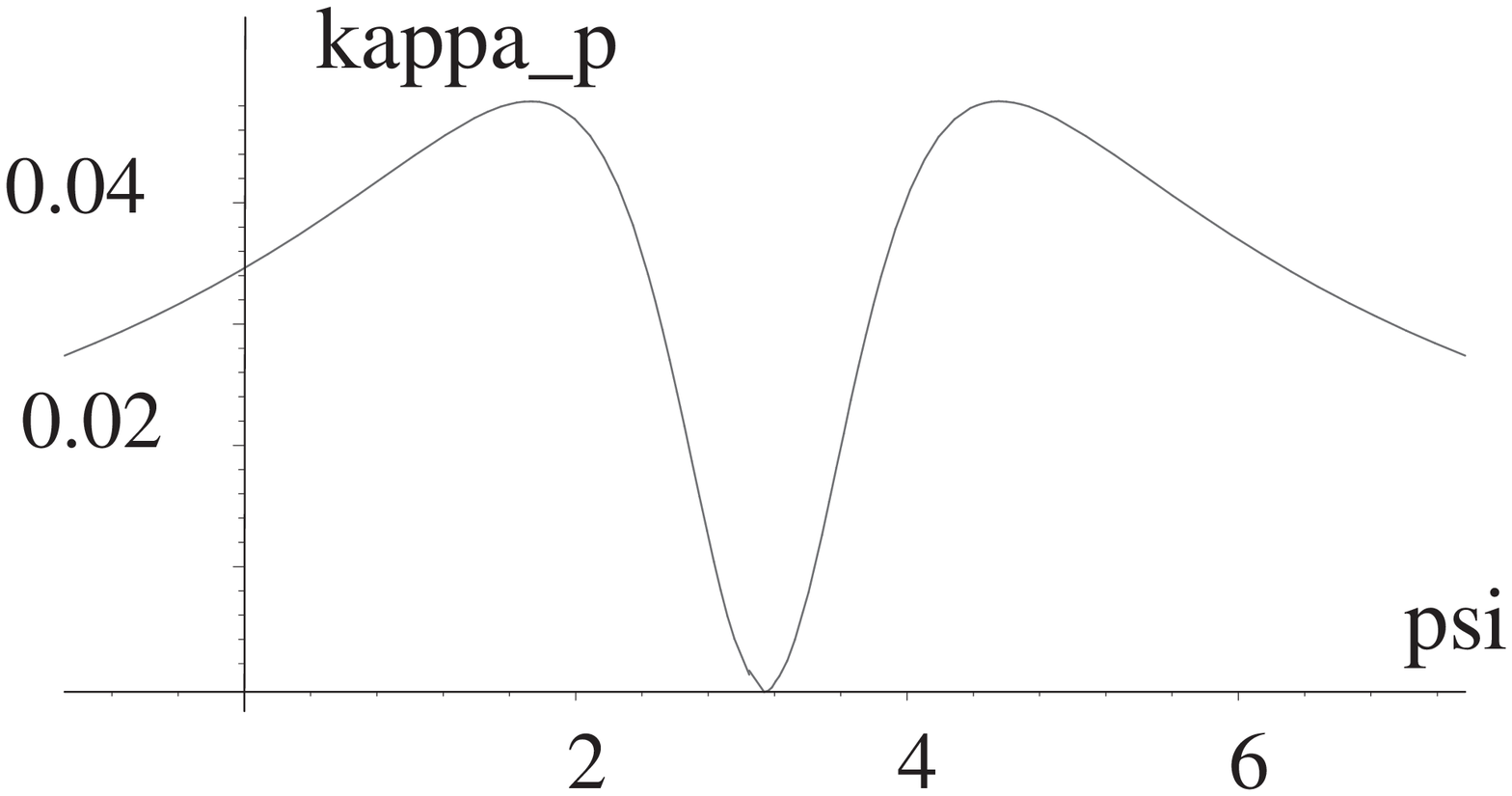,
   scale=0.17}}}
   \end{center}
 \end{minipage}
 \begin{minipage}[b]{4.45cm}
   \psfrag{kappa_p}{$\kappa_p$}
   \psfrag{psi}{$\mu$ (rad)}
   {\scriptsize
   \psfrag{0.023}{0.023}
   \psfrag{0.025}{0.025}
   \psfrag{0.027}{0.027}
   \psfrag{0}{0}
   \psfrag{4}{6}
   \psfrag{6}{6}
    \subfigure[$\eta=2/\pi$]{\psfig{file= 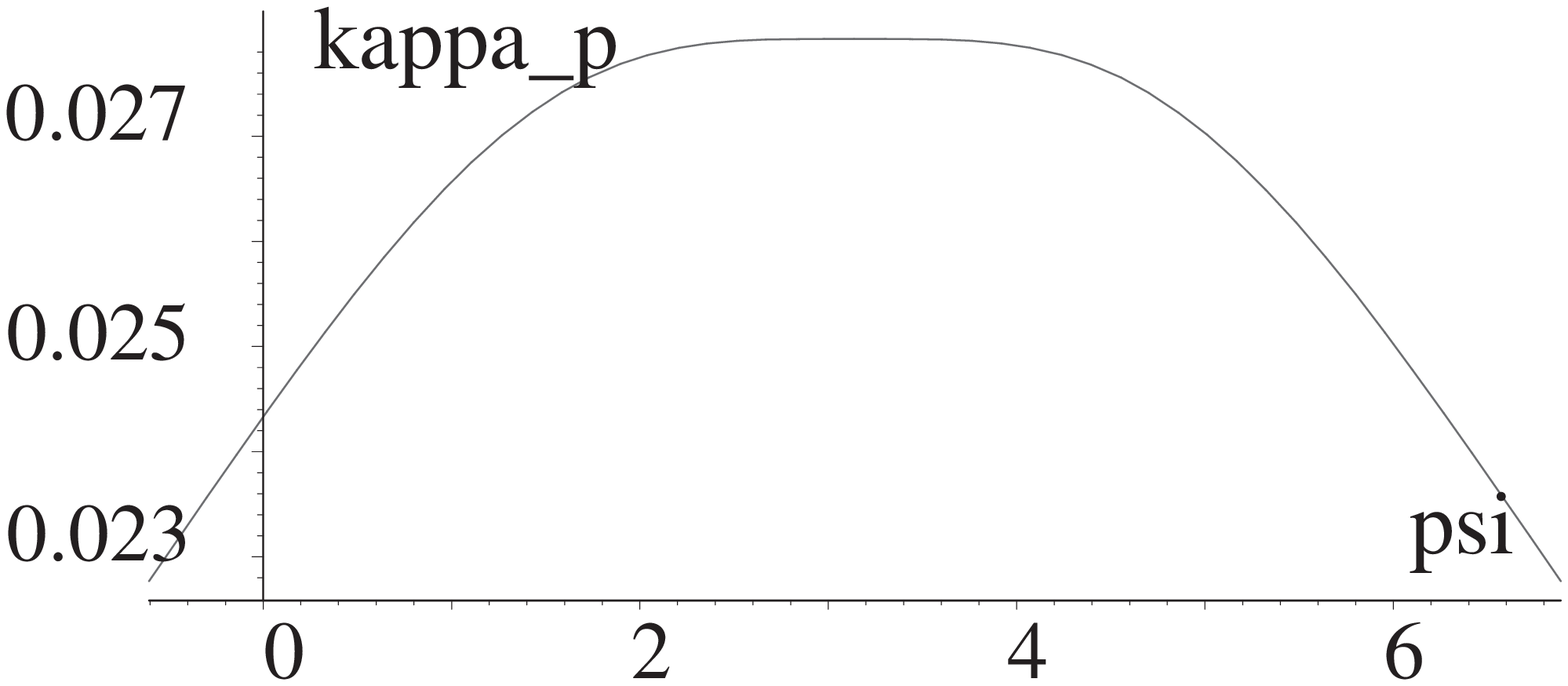,
    scale=0.17}}}
 \end{minipage}
 \caption{Pitch-curve curvature for $p=50$~mm} \label{fig101}
 \end{center}
 {\vspace{-1cm}}
\end{figure}
In summary, to have a fully convex cam profile, taking the
geometric constraints on the mechanism into consideration,
parameter $\eta$ must obey the condition given in
Eq.~(\ref{eq11}), {\it i.e.} $\eta \geq 1/\pi$. We combine the
condition on $a_{4}$ to avoid undercutting, as given in
Eq.~(\ref{eq12}), with the geometric constraints on the mechanism,
as given in Eqs.~(\ref{eq011}) and (\ref{eq012}), which are,
respectively, $a_{4} / p < 1/2$ and $a_{4} / p \leq \eta -b/p$:
 \begin{subequations}
 \begin{eqnarray}
 \mbox{if}~~ \eta \in [ \frac{1}{\pi},\frac{2}{\pi}], ~~ a_{4} &<&
\min \left\{ \frac{1}{\kappa_{p \rm max1}},~\frac{p}{2},~\eta p-b
\right\} \\
 \mbox{if}~ \eta \in [ \frac{2}{\pi},+\infty[, ~~ a_{4}
&<& \min \left\{ \frac{1}{\kappa_{p \rm max2}},~\frac{p}{2},~\eta
p-b \right\}
 \end{eqnarray}
 \label{eq21}
 \end{subequations}
where $\kappa_{p \rm max1}$ and $\kappa_{p \rm max2}$ are given in
Eqs.~(\ref{eq19}) and (\ref{eq20}), respectively.
\section{Optimization of the Roller Pin}
We concluded in previous section that the lowest values of
parameters $\eta$ and $a_{4}$ led to the lowest values of the
pressure angle. Nevertheless, we must take into consideration that
the smaller the radius of the roller, the bigger the deformation
of the roller pin, and hence, a decrease of the stiffness and the
accuracy of the mechanism. In this section we formulate and solve
an optimization problem to find the best compromise on parameters
$\eta$ and $a_{4}$ to obtain the lowest pressure angle values with
an acceptable deformation of the roller pin.
\subsection{Minimization of the Elastic Deformation on the Roller Pins}
Here we find the expression for the maximum elastic deformation on
the pin, which will be minimized under given constraints.
Figure~\ref{fig400} displays the free part of the pin, {\it i.e.},
the part not fixed to the roller-carrying slider, as a cantilever
beam, where the load $F=\sqrt{f_{x}^{2}+f_{y}^{2}}$ denotes the
magnitude of the force {\bf f} transmitted by the cam. This force
is applied at a single point at the end of the pin in the worst
loading case. Although the dimensions of the pin are not those of
a simple beam, we assume below that the pin can be modelled as
such, in order to obtain an explicit formula for its deflection.
This assumption was found to be plausible by testing it with FEA
\cite{Teng:2003}.
 \begin{figure}[htb]
 {\vspace{-0.5cm}}
  \begin{center}
   \psfrag{f}{$\bf f$}
   \psfrag{L}{$L$}
   \psfrag{a5}{$a_5$}
   \psfrag{x}{$x$}
   \psfrag{y}{$y$}
   \psfrag{z}{$z$}
   {\epsfig{file = 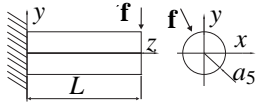,scale = 0.2}}
  \caption{Approximation of the roller pin as a cantilever beam}
  \label{fig400}
  \end{center}
 {\vspace{-1cm}}
 \end{figure}

The displacement $v_{L}$ at the free end of the pin turns out to
be
 \begin{equation}
 \label{eq400}
 v_{L}=\sqrt{v_{x}^{2}+v_{y}^{2}}=\frac{FL^{3}}{3EI}
 \end{equation}
where $E$ is the Young modulus and $I=\pi a_{5}^{4}/4$, with
$a_{5}$ denoting the radius of the pin, is the polar moment of
inertia of the cross section. Moreover, $v_{x}$ and $v_{y}$ are
the pin elastic displacements in the $x$- and $y$-directions,
respectively, at the free end.

Before proceeding, we prove that the vertical component $f_{y}$ of
the transmitted force is constant, and hence, we will consider
only the magnitude of the $x$-component of $v_{L}$. Since we
assume that the mechanism undergoes a pure-rolling motion, the
force exerted by the cam onto the roller, denoted by ${\bf
f}=[f_{x}~~f_{y}]^{T}$, passes through the center of the roller,
{\it i.e.}, its line of action passes through points $O_{2}$ and
$C$, as depicted in Fig.~\ref{fig003}. With a constant torque
$\tau$ provided by the motor, we have $\tau=||{\bf f}||d$, where
$d$ denotes the distance from the center of the input axis to the
line of action of the force {\bf f}. Moreover, we have $d=b_{2}
\sin \delta$. Hence, $\quad  \tau=||{\bf f}||(b_{2} \sin
\delta)=b_{2}(||{\bf f}|| \sin\delta)=b_{2}f_{y} $. Finally, since
$b_{2}=2\pi/p$ we obtain the expression for $f_{y}$ sought:
 \begin{equation}
  \label{eq404} f_{y}=\frac{2\pi\tau}{p}=F_{0}
 \end{equation}
Since $\tau$ is constant, $f_{y}$ is also constant throughout one
cycle. Consequently we only have to consider the $x$-component of
\negr f, and hence, $v_{x}$ for the minimization problem:
\begin{equation}
\label{eq405}
v_{x}=\frac{|f_{x}|L^{3}}{3EI}=\frac{4|f_{x}|L^{3}}{3E\pi
a_{5}^{4}}=\beta
\frac{|f_{x}|}{a_{5}^{4}},~~~\beta=\frac{4L^{3}}{3E\pi}
\end{equation}
$\beta$ thus being a constant factor. The objective function $z$,
to be minimized, is thus defined as
\begin{equation}
\label{eq406} z=\frac{f_{\rm
max}^{2}}{a_{5}^{4}}~~\rightarrow~~\min_{\eta,a_{4},a_{5}}
\end{equation}
where $f_{\rm max}$ is the maximum value of $f_{x}$ throughout one
cycle. Since
\begin{equation}
\label{eq407}
f_{x}=\frac{f_{y}}{\tan\delta}=\frac{F_{0}}{\tan\delta}
\end{equation}
we obtain
 \beqa
z=\frac{1}{a_{5}^{4}} \max_{\psi} \left\{ f_{x}^{2} \right\}
=\frac{1}{a_{5}^{4}} \max_{\psi} \left\{
\frac{F_{0}^{2}}{\tan^{2}\delta} \right \}
=\frac{F_{0}^{2}}{a_{5}^{4}} \max_{\psi} \left\{
\frac{1}{\tan^{2}\delta} \right \} \nonumber
 \eeqa
with $\delta$, a function of $\psi$, given in Eq.~(\ref{eq08}c).
Moreover, the system operates by means of two conjugate
mechanisms, which alternately take over the power transmission. We
established in Eq.~(\ref{eq0080}) that when one mechanism is in
positive action, $\psi$ is bounded between $\psi_{i}=\pi-\Delta$
and $\psi_{f}=2\pi-\Delta$, which corresponds to $\delta$ bounded
between $\delta_{i}$ and $\delta_{f}$ with $0 \leq \delta_{i} <
\delta_{f} \leq \pi$. Moreover, functions $1/\tan^{2}\delta$ and
$\cos^{2}\delta$ are both unimodal in $-\pi \leq \delta \leq 0$
and in $0 \leq \delta \leq \pi$, their common maxima finding
themselves at $-\pi$, $0$ and $\pi$. Since $\cos^{2}\delta$ is
better behaved than $1/\tan^{2}\delta$, we redefine $z$ as
 \beqa
   z=\frac{1}{a_{5}^{4}} \max_{\delta_{i} \leq \delta \leq
\delta_{f}} \left\{ \cos^{2}\delta \right \} \nonumber
  \eeqa
Furthermore, the function $\cos^{2}\delta$ attains its global
minimum of $0$ in $[0,~\pi]$, its maximum in the interval
$[\delta_{i},~\delta_{f}]$, included in $[0,~\pi]$, occurring at
the larger of the two extremes of the interval, $\delta_{i}$ or
$\delta_{f}$. It follows that the objective function to be
minimized becomes
 \beqa
  z=\frac{1}{a_{5}^{4}} \max \left\{
\cos^{2}\delta_{i},~\cos^{2}\delta_{f} \right \} \nonumber
  \eeqa
with $\delta_{i}$ and $\delta_{f}$ the values of $\delta$ for
$\psi_{i}=\pi-\Delta$ and $\psi_{f}=2\pi-\Delta$, respectively.
Using the expression for $\delta$ given in Eq.~(\ref{eq08}c) and
the trigonometric identity
\[ \cos (\arctan x)=\frac{1}{\sqrt{1+x^{2}}} \]
we obtain the expression for $\cos^{2}\delta$:
 \beqa
   \cos^{2}\delta=\frac{(2\pi \eta-1)^{2}}{(2\pi
\eta-1)^{2}+(\psi-\pi)^{2}}
  \eeqa
Hence,
 \begin{subequations}
 \begin{eqnarray}
 \cos^{2}\delta_{i} &=&
 \frac{(2\pi \eta-1)^{2}}{(2\pi\eta-1)^{2}+(\psi_{i}-\pi)^{2}}  \label{eq412}\\
 \cos^{2}\delta_{f} &=& \frac{(2\pi \eta-1)^{2}}{(2\pi\eta-1)^{2}+(\psi_{f}-\pi)^{2}}
 \end{eqnarray}
 \end{subequations}
Furthermore, since $\psi_{i}=\pi-\Delta$, $\psi_{f}=2\pi-\Delta$
and $\Delta<0$, we have $\psi_{f} > \psi_{i} > \pi$, $
\psi_{f}-\pi > \psi_{i}-\pi > 0$ and, consequently, from
Eqs.~(\ref{eq412} \& b), $\cos^{2}\delta_{i}>\cos^{2}\delta_{f}$
and the objective function to minimize becomes
 \begin{equation}
 \label{eq414}
 z=\frac{\cos^{2}\delta_{i}}{a_{5}^{4}}
 \quad  \rightarrow \quad
 \min_{\eta,a_{4},a_{5}}
\end{equation}
with $\cos^{2}\delta_{i}$ given in Eq.~(\ref{eq412}) and
$\psi_{i}=\pi-\Delta$.
\subsection{Geometric Constraints}
Two neighboring pins cannot be tangent to each other, as depicted
in Fig.~\ref{fig401}, and hence the radius $a_{5}$ of the pin is
bounded as
 \be
    a_{5} / p < 1/4
 \label{eq4140}
 \ee
Furthermore, $a_{4}$ and $a_{5}$ are not independent. From the SKF
catalogue, for example, we have information on bearings available
in terms of the outer radius $D$ and the inner radius $d$, as
shown in Fig.~\ref{fig402}. We divide these bearings into five
series, from 1 to 5. Hence, each series can be represented by a
continuous function. We chose series 2, in which the basic dynamic
load rating $C$ lies between 844 and $7020$~N. Furthermore, for
series 2, the relation between $D$ and $d$ can be approximated by
a linear function $D$ vs. $d$, $D \simeq 1.6d+10$ (in mm). Since
$D=2a_{4}$ and $d=2a_{5}$, the above equation leads to
\begin{equation}
\label{eq416} a_{4} \simeq 1.6 a_{5}+5 ~~~\mbox{in mm}
\end{equation}
 \begin{figure}
  \begin{center}
  \begin{minipage}[b]{3cm}
   \begin{center}
   {\scriptsize
   \psfrag{p}{$p$}
   \psfrag{p/2}{$p/2$}
   \psfrag{a4}{$a_4$}
   \psfrag{a5}{$a_5$}
   {\epsfig{file =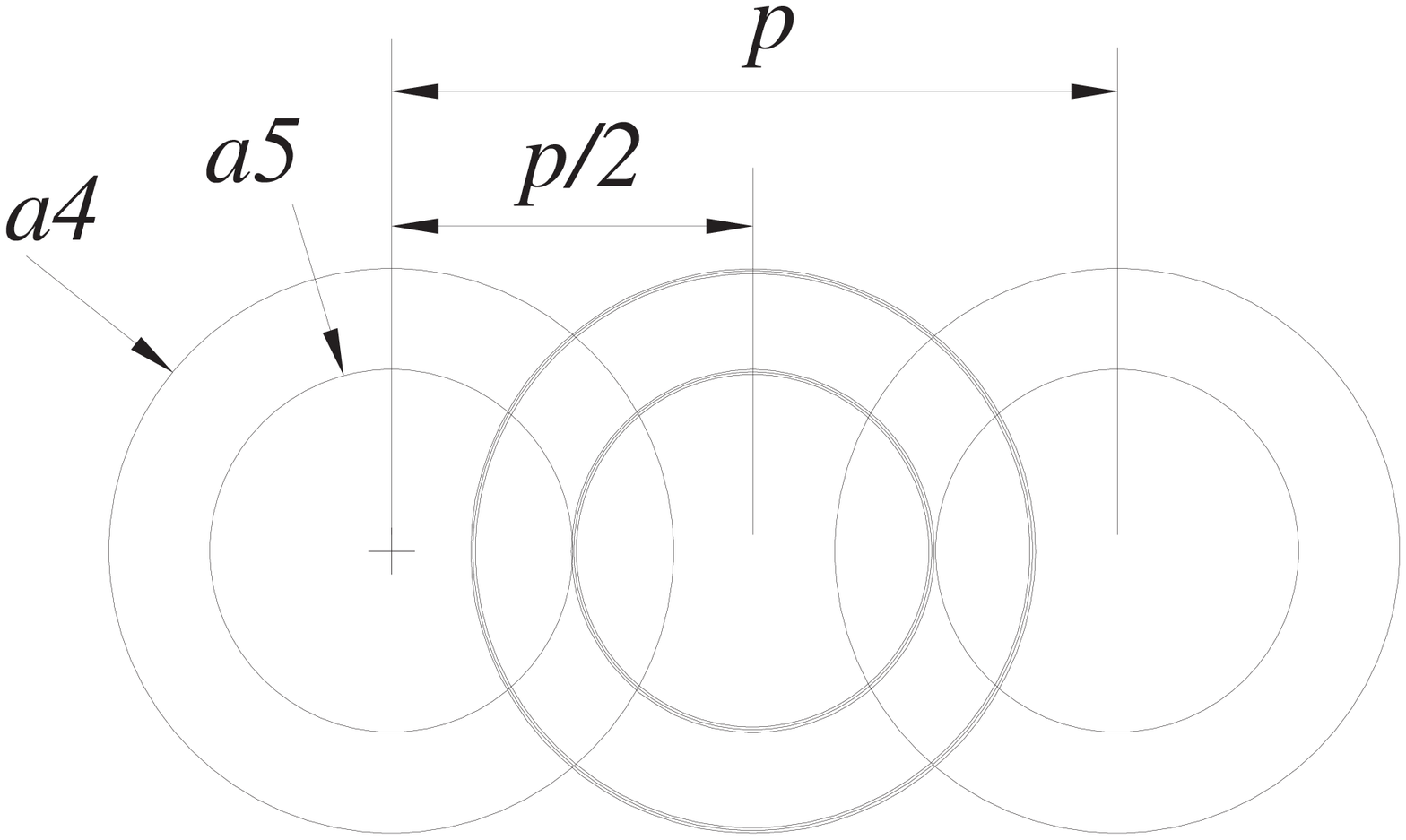,scale = 0.11}}}
   \caption{Geometric constraint on the roller-pin radius}
   \label{fig401}
   \end{center}
  \end{minipage}
  \begin{minipage}[b]{5.5cm}
   {\scriptsize
  \psfrag{10}{10}\psfrag{20}{20}\psfrag{30}{30}\psfrag{40}{40}
  \psfrag{50}{50}\psfrag{60}{60}\psfrag{70}{70}\psfrag{0}{0}
  \psfrag{15}{15}\psfrag{5}{5}
  \psfrag{Series 1}{Series 1}   \psfrag{Series 2}{Series 2}   \psfrag{Series 3}{Series 3}
  \psfrag{Series 4}{Series 4}   \psfrag{Series 5}{Series 5}
  \psfrag{d (mm)}{$d$ (mm)}     \psfrag{D (mm)}{$D$ (mm)}}
  \psfig{file= 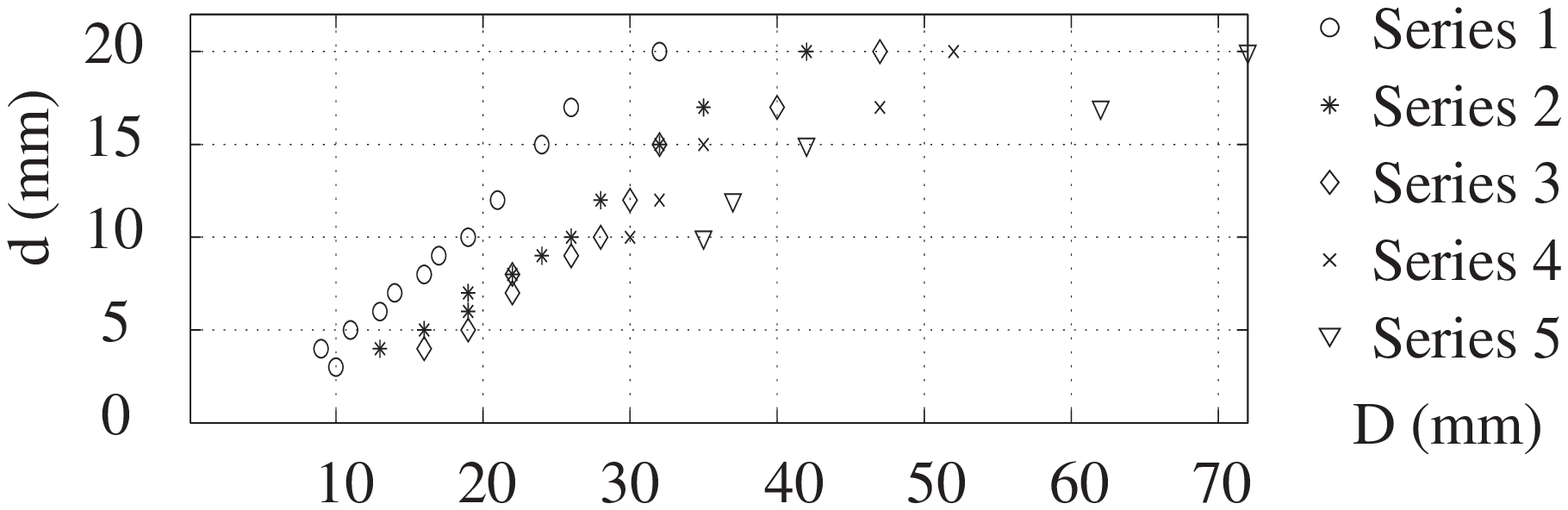, scale= 0.3}
  \caption{Dimensions of the SKF bearings}
  \label{fig402}
  \end{minipage}
   \end{center}
 {\vspace{-0.5cm}}
 \end{figure}

\noindent We define now two non-dimensional parameters
$\alpha_{4}$ and $\alpha_{5}$:
 \begin{equation}
 \label{eq417}
 \alpha_{4}= a_{4}/p \quad~~\alpha_{5}=a_{5}/ p
 \end{equation}
where $\alpha_{5}$ can be derived from Eq.~(\ref{eq416}) as
 \begin{equation}
 \label{eq418} \alpha_{5}=(5/8)\alpha_{4}-25/(8p)
 \quad \mbox{with $p$ in mm.}
 \end{equation}
The optimization problem is now expressed as
 \begin{equation}
 \label{eq419}
z(\eta,\alpha_{4})=\frac{\cos^{2}\delta_{i}}{\alpha_{5}^{4}}~~\rightarrow~~\min_{\eta,\alpha_{4}}
 \end{equation}
Moreover, we recall the geometric constraint defined in
Eqs.~(\ref{eq11}) and (\ref{eq21}) that can be rewritten as
\begin{subequations}
\begin{eqnarray}
 g_{1}&=& \frac{1}{\pi}-\eta \leq 0  \\
 g_{2}&=& \alpha_{4}-\frac{1}{2} < 0 \\
 g_{3}&=& \alpha_{4}-\frac{1}{p\kappa_{p \rm max}} < 0 \\
 g_{4}&=& \alpha_{4}-\eta+\frac{b}{p} \leq 0 \\
 g_{5}&=& \alpha_{5}-\frac{1}{4} <0
\end{eqnarray}
\label{eq420}
\end{subequations}
with $\cos^{2}\delta_{i}$ and $\alpha_{5}$ given in
Eqs.~(\ref{eq412}) and (\ref{eq418}), respectively.
\subsection{Results of the Optimization Problem}
We solve the foregoing optimization problem with an algorithm
implemented Matlab, using $p=50$~mm and $b=9.5$~mm. One solution
is found, corresponding to $\eta=0.69$ and $a_{4}=24.9992$~mm,
with a value of $z=249$. The algorithm finds the values of $\eta$
and $a_{4}$ as big as possible considering the constraints. For
this solution, constraints (\ref{eq420}d \& e) are active.
Nevertheless, as we saw in subsection 3.3. about the influence of
parameters $h$ and $a_{4}$, we want these parameters to be as
small as possible in order to have low pressure-angle values. For
the solution found above, the service factor is equal to 0\%.
Consequently, we must find a compromise between the pressure angle
and the roller-pin elastic deformation. Table~\ref{tab401} shows
solutions found by the optimization algorithm upon reducing the
boundaries of $\eta$. Each time the algorithm finds the
corresponding value of $a_{4}$ as big as possible, constraint
(\ref{eq420}d) becomes active. Recorded in this table is also the
corresponding maximum elastic deformation of the roller pin $v_{L
\rm max}$ (its expression is derived below), the minimum and the
maximum absolute values of the pressure angle, $|\mu_{\rm min}|$
and $|\mu_{\rm max}|$, respectively, and the service factor, as
defined in section 2.4. From Eqs.~(\ref{eq400}) and (\ref{eq405})
we have
  \beqa
    v_{L}=\frac{\beta}{a_{5}^{4}}\sqrt{f_{x}^{2}+f_{y}^{2}}
    \nonumber
  \eeqa
Using Eqs.~(\ref{eq404}) and (\ref{eq407}), the above equation
leads to
 \beqa
 v_{L}=\frac{\beta
 F_{0}}{a_{5}^{4}}\sqrt{1+\frac{1}{\tan^{2}\delta}}     \nonumber
 \eeqa
Which can be simplified by means of the expression for
$\cos^{2}\delta_{i}$ given in Eq.~(\ref{eq412}) as
 \begin{equation}
 \label{eq423}
 v_{L\rm max}=\frac{\beta F_{0}}{a_{5}^{4}} \frac
{\sqrt{(2\pi \eta-1)^{2}+(\psi_{i}-\pi)^{2}}}{|\psi_{i}-\pi|}
 \end{equation}
with $\beta$, $F_{0}$ and $a_{5}$ given in Eqs.~(\ref{eq405}),
(\ref{eq404}) and (\ref{eq416}), respectively, and
$\psi_{i}=\pi-\Delta$. In Table~\ref{tab401} we record the value
of $v_{L\rm max}$ with $L=10$~mm, $\tau=1.2$~Nm (according to the
Orthoglide specifications recalled in section 1) and $E=2 \times
10^{5}$~MPa. We conclude from Table~\ref{tab401} that for this cam
profile we cannot find an acceptable compromise between a low
deformation of the roller pin, and hence a high stiffness and
accuracy of the mechanism, and low pressure-angle values. Indeed,
for an acceptable deformation of the roller pin, $v_{L {\rm
max}}=8.87~\mu$m, obtained with $\eta=0.38$, the service factor
equals 54.68\%, which is too low. On the other hand, for an
acceptable service factor of 79.43\%, obtained with $\eta=1/\pi$,
the deformation of the roller pin is equal to 710.19~$\mu$m.
\begin{table}
\begin{center}
{\scriptsize
\begin{tabular}{|c|m{0.4cm}|m{0.5cm}|m{0.9cm}|m{0.5cm}|m{0.7cm}|m{0.8cm}|m{1cm}|}
\hline $\eta$  &  $a_{4}$ (mm)  &  $a_{5}$ (mm)  &  $z$ & $v_{L\rm
max}$ ($\mu$m)  &  $|\mu_{\rm min}|$ ($^{\circ}$)  &  $|\mu_{\rm
max}|$ ($^{\circ}$)  &service factor (\%)\\
 \hline
 0.69 & 24.99 & 12.50 & 249 & 0.09   & 42.11 & 80.68 & 0 \\
 0.5 & 15.5 & 6.56 & 2968 & 0.50 & 28.59 & 69.81 & 6.85 \\
 0.4 & 10.5 & 3.44 & 32183 & 4.32  & 20.31 & 57.99 & 46.68 \\
 0.39 & 10 & 3.12 & 45490 & 6.07 & 19.46 & 56.42 & 50.68 \\
 0.38 & 9.5 & 2.81 & 66659 & 8.87 & 18.61 & 54.78 & 54.68 \\
 0.37 & 9 & 2.50 & 102171 & 13.63 & 17.75 & 53.04 & 58.69 \\
 0.36 & 8.5 & 2.19 & 165896 & 22.31  & 16.89 & 51.22 & 62.69\\
 0.35 & 8 & 1.87 & 290765 & 39.71 & 16.03 & 49.31 & 66.70\\
 0.34 & 7.5 & 1.56 & 566521 & 79.18 & 15.17 & 47.31 & 70.72 \\
 0.33 & 7 & 1.25 & 1.29 10$^{6}$ & 186.06  & 14.31 & 45.21 & 74.73 \\
 $1/\pi$ & 6.41 & 0.88 & 4.68 10$^{6}$ & 710.19  & 13.31 & 42.64 & 79.43\\
 \hline
\end{tabular}
}
 \end{center}
 \caption{Results of the optimization problem, with
$p=50$~mm, $b=9.5$~mm, $L=10$~mm, $\tau=1.2$~Nm and $E=2 \times
10^{5}$~MPa}
 \label{tab401}
\end{table}
 {\vspace{-1cm}}
\section{A Non-Coaxial Conjugate-Cam Mechanism}
This Section describes a new mechanism, based on Slide-O-Cam, that
enables us to decrease considerably the pressure angle while
meeting the Orthoglide specifications. This mechanism is composed
of three conjugate cams mounted on three parallel shafts, the
rollers being placed on one single side of the roller-carrying
slider. One motor provides the torque to the central camshaft,
this torque then being transmitted to the two other camshafts
through a parallelogram mechanism coupling them, whose detailed
design is reported in \cite{Renotte:2003}. We denote by 1, 2 and 3
the three cams, as shown in Fig.~\ref{fig600}.
\begin{figure}
 {\vspace{-0.5cm}}
 \center
{\scriptsize
 \psfrag{p}{$p$}  \psfrag{p/2}{$p/2$}
 \psfrag{1}{1}  \psfrag{2}{2}  \psfrag{3}{3}
 \psfrag{O1}{$O_1$}
 \psfrag{xu}{$x, u$}  \psfrag{yv}{$y, v$}
 \psfrag{u}{$u$}  \psfrag{v}{$v$}
 \psfrag{s(2Pi/3)}{$s(2 \pi / 3)$}  \psfrag{s(4Pi/3)}{$s(4 \pi / 3)$}
 \psfrag{y12}{$y_{12}$} \psfrag{y13}{$y_{13}$}
 \psfig{file=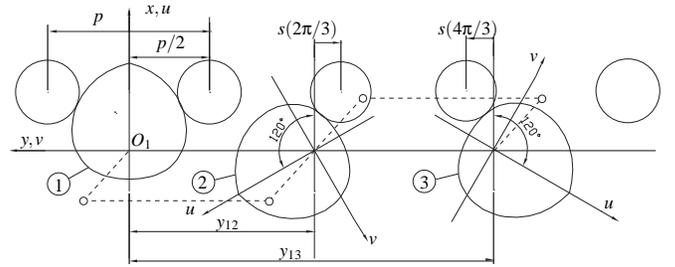, scale=0.4}}
 \caption{Layout of the non-coaxial conjugate-cam mechanism}
\label{fig600}
 {\vspace{-0.5cm}}
\end{figure}
The profile of each cam is described in Section 2. The cams are
mounted in such a way that the angle between the $u$-axis of cam 1
and cam 2 is 120$^{\circ}$, and the angle between the $u$-axis of
cam 1 and cam 3 is 240$^{\circ}$. According to the configuration
of the mechanism depicted in Fig.~\ref{fig600}, and denoting by
$y_{12}$ and $y_{13}$ the distance between the origin of 1 and 2,
and between the origin of 1 and 3, respectively, we have
 \beqa
 y_{12} = p/2+p+s(2\pi/3)~,~~~ y_{13} = p/2 + 2p + s(4\pi/3) \nonumber
 \eeqa
Using the expression of the input-output function $s$ given in
Eq.~(\ref{eq01}), we obtain
 \be
 \label{eq601} y_{12} = 4p/3~,~~~ y_{13} = 8p/3
 \ee
Figure~\ref{fig601} shows the pressure angle variation for each
cam vs.\ $\psi$ where \textit{1}, \textit{2} and \textit{3} denote
the plot of the pressure angle for cams 1, 2 and 3, respectively.
Moreover, cams \textit{2} and \textit{3} are rotated by angles
$2\pi /3$ and $4\pi/3$, respectively, from cam \textit{1}. We can
also consider that the plot for cam 3 that drives the follower
before cam 1 in a previous cycle, refereed to as \textit{3'} in
Fig.~\ref{fig601}, is a translation of $-2\pi/3$ from cam
\textit{1}. As we saw in Eq.~(\ref{eq01111}), cam 1 can drive the
follower within $\pi \leq \psi \leq 2\pi -\Delta$, which
corresponds in Fig.~\ref{fig601} to the part of the plot
\textit{1} between points $B$ and $D$. Consequently, cam 2 can
drive the follower within
 {\vspace{-0.1cm}}
 \begin{eqnarray*}
 \pi+ 2\pi/3 \leq &~\psi~& \leq 2\pi -\Delta+ 2\pi/3
 \quad i.e. \quad
 5\pi / 3 \leq \psi \leq 8\pi/3-\Delta
 \end{eqnarray*}
 {\vspace{-0.1cm}}
and cam 3 within
 {\vspace{-0.1cm}}
 \begin{eqnarray*}
 \pi+ 4\pi/3 \leq &~\psi~& \leq 2\pi -\Delta+ 4\pi/3
 \quad i.e. \quad
 7\pi/3 \leq \psi \leq 10\pi/3-\Delta
 \end{eqnarray*}
 {\vspace{-0.1cm}}
which is equivalent to saying that cam 3 can drive the follower in
a previous cycle, within
 {\vspace{-0.1cm}}
\begin{eqnarray*}
 \pi-2\pi/3 \leq &~\psi~& \leq 2\pi -\Delta-2\pi/3
 \quad i.e. \quad
 \pi/3 \leq \psi \leq 4\pi/3-\Delta
\end{eqnarray*}
 {\vspace{-0.1cm}}
The above interval corresponds in Fig.~\ref{fig601} to the part of
the plot \textit{3'} between points $A$ and $C$. Consequently,
there is a common part for cams 1 (plot \textit{1}) and 3 (plot
\textit{3'}) during which these two cams can drive the follower,
namely, between points $B$ and $C$, which corresponds to
$
 \pi \leq \psi \leq 4\pi/3-\Delta \label{eq602}$.
Moreover, during this common part, cam 3 has lower absolute
pressure angle values than 1, and hence, we consider that only cam
3 drives the follower. Consequently, cam 1 drives the follower
only within $4\pi/3-\Delta \leq \psi \leq 2\pi-\Delta
\label{eq603}$. These boundaries allow us to have a pressure angle
lower than with coaxial conjugate cams, since we do not use
anymore the part of the cam profile within $\pi-\Delta \leq \psi
\leq 4\pi/3-\Delta$, which has high absolute pressure angle
values. We can thus obtain a higher service factor for the
mechanism. The drive thus design is sketched in Fig.~\ref{fig700}.
 \begin{figure}[!hb]
 {\vspace{-0.5cm}}
 \begin{minipage}[b]{5.0cm}
 \center
 \psfrag{mu}{$\mu$}
 \psfrag{3p}{\textit{3'}} \psfrag{1}{\textit{1}} \psfrag{2p}{\textit{2}} \psfrag{3}{\textit{3}}
{\scriptsize
 \psfrag{20}{20}   \psfrag{40}{40}   \psfrag{60}{60}   \psfrag{80}{80}   \psfrag{100}{100}
 \psfrag{-20}{-20} \psfrag{-40}{-40} \psfrag{-60}{-60} \psfrag{-80}{-80}
 \psfrag{A}{$A$}  \psfrag{B}{$B$}  \psfrag{C}{$C$}  \psfrag{D}{$D$}
 \psfrag{-2}{-2} \psfrag{-4}{-4}
 \psfrag{2}{2} \psfrag{4}{4}  \psfrag{6}{6}  \psfrag{8}{8} \psfrag{10}{10}
 \psfig{file=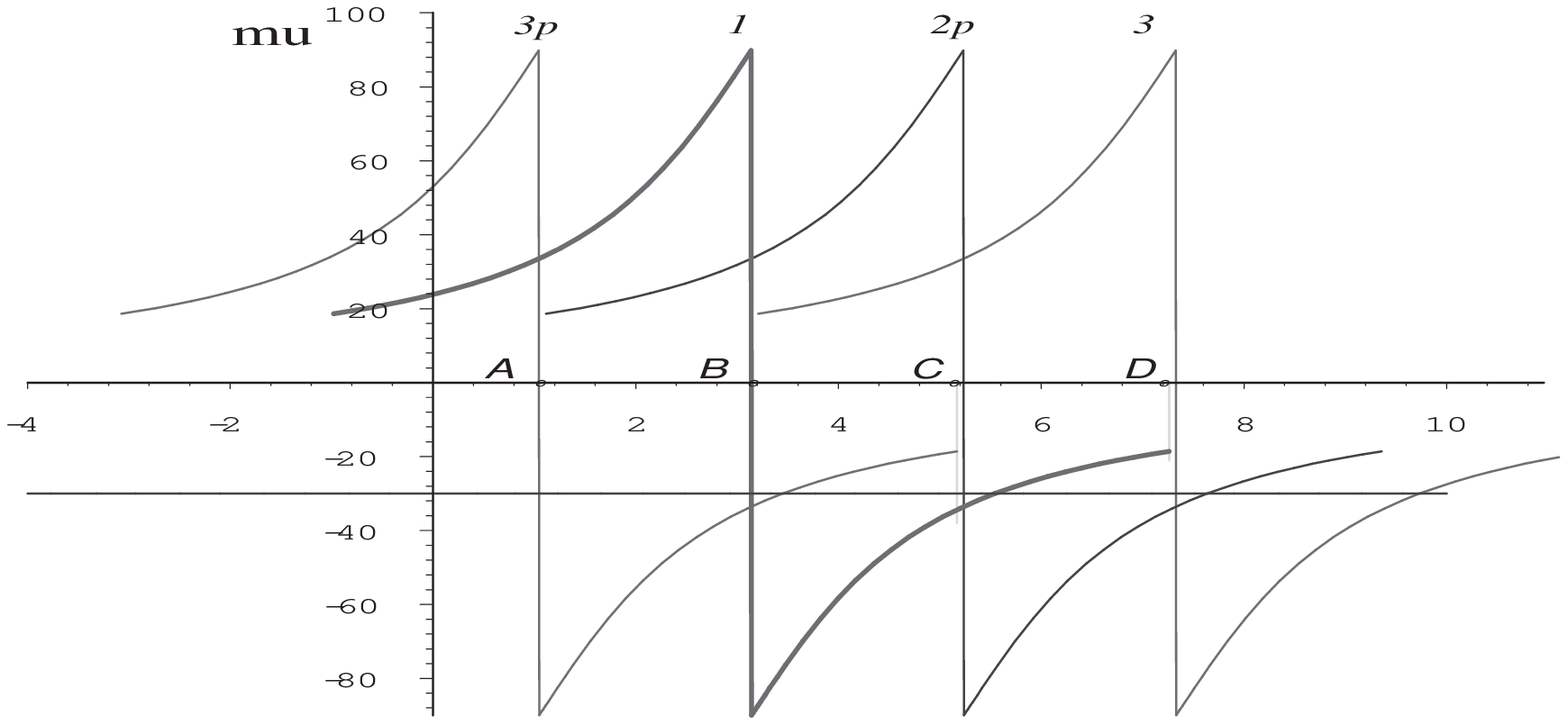, scale=0.3}}
 \caption{Pressure angle for the three cams} \label{fig601}
 \end{minipage}
 \begin{minipage}[b]{3.5cm}
 \center
{\scriptsize
 \psfrag{-15}{-15} \psfrag{-10}{-10}   \psfrag{-5}{-5}   \psfrag{5}{5}   \psfrag{10}{10}   \psfrag{15}{15}   \psfrag{20}{20}
 \psfig{file=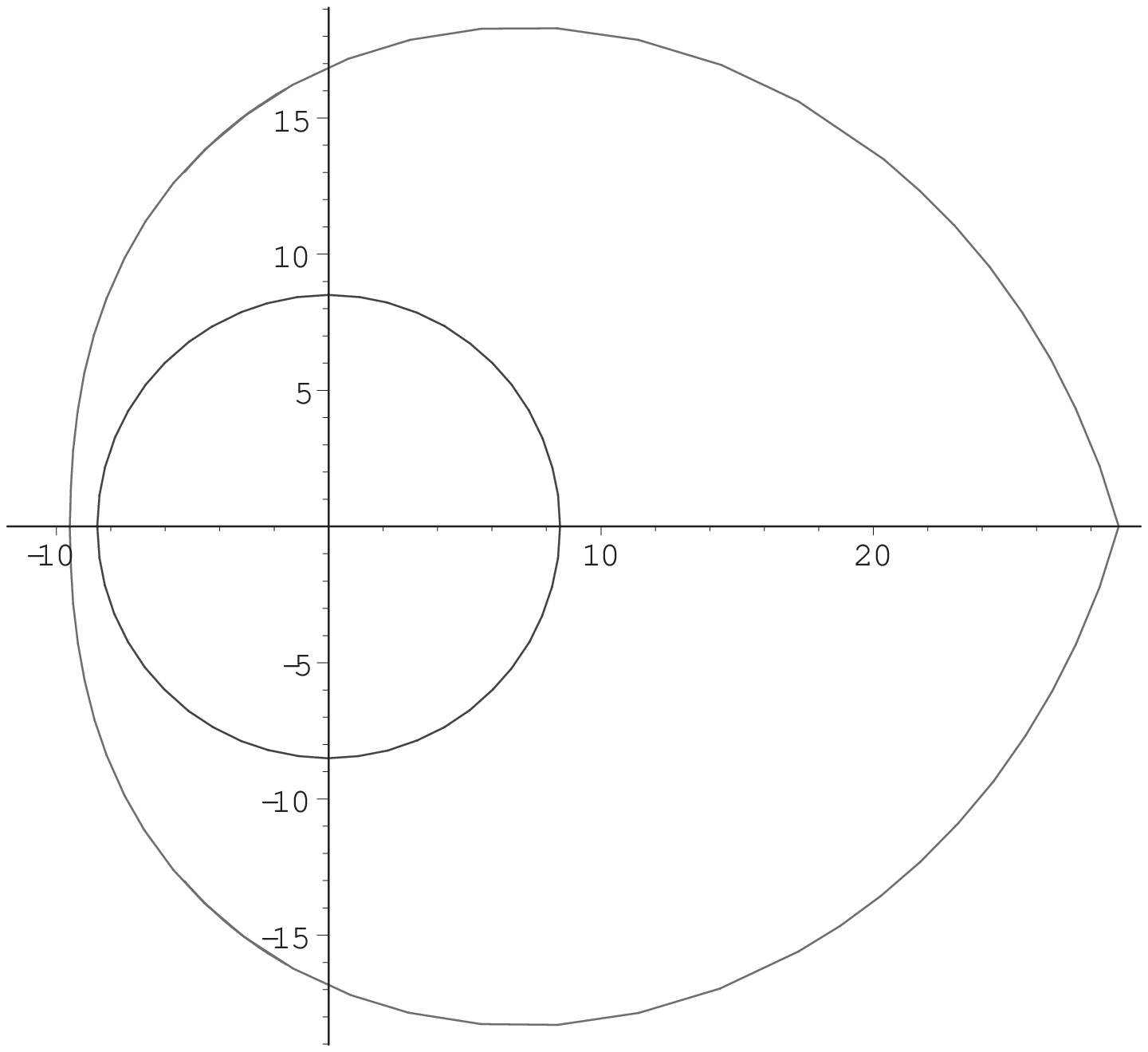, scale=0.2}}
 \caption{A sketch of the prismatic drive}
 \label{fig700}
 \end{minipage}
 {\vspace{-0.5cm}}
\end{figure}
 {\vspace{-1cm}}
\section{Conclusions}
The non-coaxial conjugate-cam mechanism reported here allows us to
drive the Orthoglide with prismatic actuators using rotary DC
motors. Moreover, the maximum roller-pin deformation $v_{L \rm
max}$ derived in eq.(\ref{eq423}) is reasonable low. In
Table~\ref{tab600}, we record the values of $\eta$, $a_{4}$,
$a_{5}$, $v_{L \rm max}$, $|\mu_{\rm min}|$, $|\mu_{\rm max}|$ and
the service factor for the non-coaxial conjugate-cam mechanism.
The best compromise is to use the non-coaxial conjugate-cam
mechanism with $\eta=0.37$, whence the radius of the roller is
$a_{4}=9$~mm and the roller-pin deformation is $v_{L \rm
max}=9.76~\mu$m with a good service factor of 88.03\%.
\begin{table}
\begin{center}
{\scriptsize
\begin{tabular}{|m{0.3cm}|m{0.9cm}|m{0.9cm}|m{0.9cm}|m{0.9cm}|m{0.9cm}|m{1cm}|}
\hline $\eta$  &  $a_{4}$ (mm)  &  $a_{5}$ (mm)  & $v_{L \rm max}$
($\mu$m)  & $|\mu_{\rm min}|$ ($^{\circ}$)  &  $|\mu_{\rm max}|$
($^{\circ}$)  &service factor (\%)\\  \hline
 0.5 & 15.5 & 6.56 & 0.26 & 28.59 & 49.41 & 10.49 \\
 0.4 & 10.5 & 3.44  & 2.88 & 20.31 & 37.20 & 70.02 \\
 0.39 & 10 & 3.12 & 4.14 & 19.46 & 35.81 & 76.02 \\
 0.38 & 9.5 & 2.81 & 6.20 & 18.61 & 34.39 & 82.02 \\
 0.37 & 9 & 2.50 & 9.76 & 17.75 & 32.95 & 88.03 \\
 0.36 & 8.5 & 2.19  & 16.39 & 16.89 & 31.48 & 94.04 \\
 0.35 & 8 & 1.87  & 29.89 & 16.03 & 29.98 & 100 \\
 0.34 & 7.5 & 1.56  & 61.07  & 15.17 & 28.47 & 100 \\
 0.33 & 7 & 1.25  & 147.02 & 14.31 & 26.93 & 100 \\
 $1/\pi$ & 6.41 & 0.88  & 576.95 & 13.31 & 25.12 & 100 \\ \hline
\end{tabular}}
\end{center}
\caption{Design parameters, roller pin deformation and pressure
angle for the non-coaxial conjugate-cam mechanism, with $p=50$~mm,
$b=9.5$~mm, $L=10$~mm, $\tau=1.2$~Nm and $E=2 \times 10^{5}$~MPa.}
\label{tab600} {\vspace{-0.5cm}}
\end{table}
\bibliographystyle{unsrt}

\begin{thebibliography}{99}
{\small
\bibitem{Gonzalez-Palacios:2000}
[1]~Gonz\'alez-Palacios, M.\ A. and Angeles, J., 2000, ``The novel
design of a pure-rolling transmission to convert rotational into
translational motion,'' Proc. 2000 ASME Design Engineering
Technical Conferences, Baltimore, Sept. 10-13, CD-ROM.
\bibitem{Chablat:2003}
[2]~Chablat, D. and Wenger Ph.\, 2003, ``Architecture Optimization
of a 3-DOF Parallel Mechanism for Machining Applications, the
Orthoglide,'' IEEE Transactions on Robotics and Automation, Vol.
19/3, pp. 403-410, June.
\bibitem{Lampinen:2003}
[3]~Lampinen, J., 2003, ``Cam shape optimisation by genetic
algorithm,'' Computer-Aided Design, Vol.~35, pp.~727–-737.
\bibitem{Bouzakis:1997}
[4]~Bouzakis, K.D., Mitsi, S. and Tsiafis, J., 1997,
``Computer-Aided Optimum Design and NC Milling of Planar Cam
Mechanims,'' Int. J. Math. Tools Manufaet. Vol. 37, No. 8, pp.
1131-1142.
\bibitem{Gonzalez-Palacios:1993}
[5]~Gonz\'ales-Palacios, M. A. and Angeles, J., 1993, \textit{Cam
Synthesis}, Kluwer Academic Publishers B.V., Dordrecht.
\bibitem{Waldron:1999} 
[6]~Waldron, K.\ J.\ and Kinzel, G.\ L., 1999, {\em Kinematics,
Dynamics, and Design of Machinery}, John Wiley \& Sons, Inc., New
York.
\bibitem{Lee:2001}
[7]~Lee, M.K., 2001, \textit{Design for Manufacturability of
Speed-Reduction Cam Mechanisms}, M.Eng.\ Thesis, Dept.\ of
Mechanical Engineering, McGill University, Montreal.
\bibitem{Angeles:1991}
[8]~Angeles, J. and L\'opez-Caj\'un,C., 1991, \textit{Optimization
of Cam Mechanisms}, Kluwer Academic Publishers B.V., Dordrecht.
\bibitem{Teng:2003}
[9]~Teng, C.P. and Angeles, J., 2004, ``An optimality criterion
for the structural optimization of machine element,'' ASME Journal
of Mechanical Design, Accepted for publication.
\bibitem{Renotte:2003} 
[10]~Renotte, J.\ and Angeles, J.\, ``The Design of a Pure-Rolling
Ca Mechanism to Convert Rotationnal into Translational Motion,''
Internal Report, TR-CIM-03-06, August, 2003.}
\end{thebibliography}

\end{document}